\documentclass[letterpaper, journal]{IEEEtran}  
\IEEEoverridecommandlockouts                              

\usepackage[english]{babel}
\usepackage[T1]{fontenc}

\usepackage[nolist]{acronym}
\usepackage{algorithm}
\usepackage{algpseudocode}
\usepackage[cmex10]{amsmath}
\usepackage{amssymb}
\usepackage{bm}
\usepackage{booktabs}
\usepackage{color}
\usepackage{float}
\usepackage{graphicx}
\usepackage{hyperref}
\usepackage{import}
\usepackage{mathtools}
\usepackage{multirow}
\usepackage{outline}
\usepackage{paralist}
\usepackage{siunitx}
\usepackage{stmaryrd}
\usepackage{subfigure}
\usepackage{systeme}
\usepackage{units}
\usepackage{url}
\usepackage{tikz}
\usepackage{booktabs}
\usetikzlibrary{tikzmark}

\newcommand{\mnorm}[1]{\left\| #1 \right\|}

\newcommand{\mmat}[1]{\begin{bmatrix} #1 \end{bmatrix} }

\newcommand{\figref}[1]{Fig.~\ref{#1}}
\newcommand{\secref}[1]{Section~\ref{#1}}

\newcommand{\mmod}[1]{{\color{black} #1}}

\title{\LARGE \bf Perception-aware receding horizon trajectory planning \\ for multicopters with visual-inertial odometry}
\author{Xiangyu Wu$^1$, Shuxiao Chen$^2$, Koushil Sreenath$^2$, and Mark W.\ Mueller$^1$
\thanks{$^1$Authors are with the High Performance Robotics Laboratory (HiPeRLab), UC Berkeley. {\tt\small \{wuxiangyu,mwm\}@berkeley.edu}}
\thanks{$^2$Authors are with the Hybrid Robotics Group, UC Berkeley. {\tt\small \{shuxiao.chen,koushils\}@berkeley.edu}}
}
\begin{document}
\maketitle

\begin{abstract}
Visual inertial odometry (VIO) is widely used for the state estimation of multicopters, but it may function poorly in environments with few visual features or in overly aggressive flights.
In this work, we propose a perception-aware collision avoidance \mmod{trajectory planner} for multicopters, that may be used with any feature-based VIO algorithm.
Our approach is able to fly the vehicle to a goal position at high speed, avoiding obstacles in an \mmod{unknown stationary environment} while achieving good VIO state estimation accuracy.
The proposed planner samples a group of minimum jerk trajectories and finds collision-free trajectories among them, which are then evaluated based on their speed to the goal and perception quality.
Both the motion blur of features and their locations are considered for the perception quality.
\mmod{Our novel consideration of the motion blur of features enables automatic adaptation of the trajectory’s aggressiveness under environments with different light levels.}
The best trajectory from the evaluation is tracked by the vehicle and is updated in a receding horizon manner when new images are received from the camera.
Only generic assumptions about the VIO are made, so that the planner may be used with various existing systems.
The proposed method can run in real time on a small embedded computer on board. We validated the effectiveness of our proposed approach through experiments in \mmod{both} indoor and outdoor environments.
Compared to a perception-agnostic planner, the proposed planner kept more features in the camera's view and made the flight less aggressive, making the VIO more accurate.
It also reduced VIO failures, which occurred for the perception-agnostic planner but not for the proposed planner.
\mmod{The ability of the proposed planner to fly through dense obstacles was also validated.}
The experiment video can be found at \url{https://youtu.be/qO3LZIrpwtQ}.
\end{abstract}

\section{Introduction}
\label{sec:introduction}
Multicopters are useful for a wide range of applications such as aerial photography \cite{photography} inspection \cite{inspection}, \mmod{search and rescue \cite{ceberus}}, and transportation \cite{wu2021model} thanks to their simple design and high maneuverability.
\mmod{
State estimation is necessary for these applications, which often use  onboard sensors such as the GPS \cite{GPS-flight}, camera \cite{geoloc-flight}, Lidar \cite{lidar-flight}, and inertial measurement unit \cite{imu-flight}.
For indoor applications where special localization infrastructures can be deployed, a motion capture system \cite{mocap-flight} or an  ultra-wideband system \cite{uwb-flight} could also be used.
Among state estimation methods,} visual inertial odometry (VIO) is a popular solution: it only requires light-weight, low-power, and low-cost onboard sensors -- cameras and inertial measurement units (IMUs), suitable even for small aerial robots \cite{aerialVIOoverview}. 
Additionally, VIO does not require other infrastructure in the operating environment.
These advantages make it especially useful in applications where the GPS signal is unreliable, such as indoors, in the forest, or near tall buildings.
\par
However, VIO may struggle when the vehicle flies in areas with few visual features or when the motion of the vehicle becomes too aggressive. 
As a result, the trajectory planning of a vehicle should include perception-awareness: it should consider not only the goal of the mission, but also the trajectory's impact on the VIO. 
This topic has drawn increased research interest over the past few years \cite{drone_review}. 

\begin{figure}[!t]
    \begin{center}
    \includegraphics[width=\columnwidth]{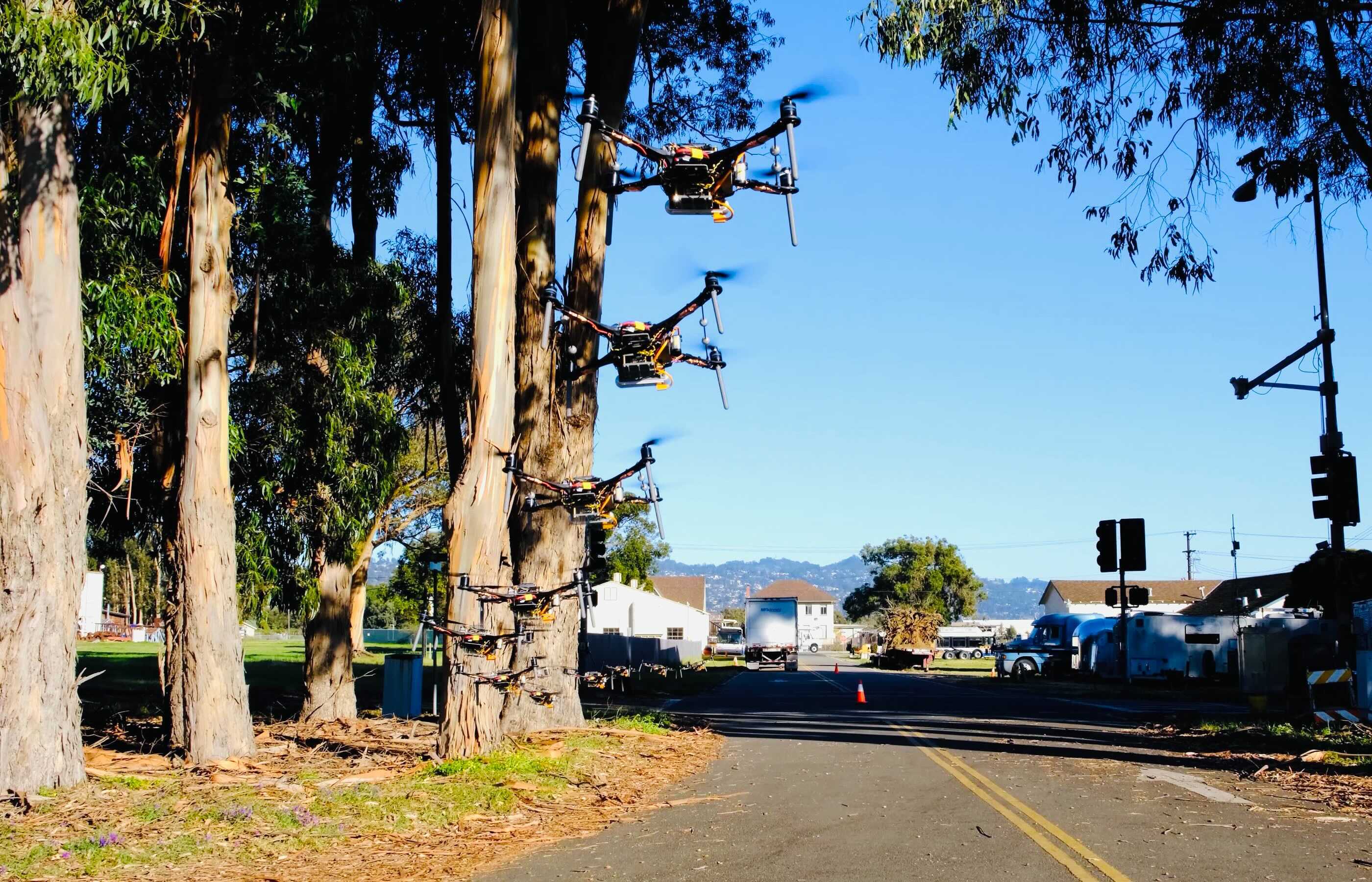}
    \end{center}
    \caption{The perception-aware planner guides the quadcopter to fly close to areas with more visual features for better VIO accuracy -- the paved road has much fewer visual features than the trees. \mmod{The goal is 20 meters forward from the starting position of the vehicle, which is in the direction away from the camera taking this photo.}}
    \label{fig:outdoor_exp}
\end{figure}

\par
\mmod{One major category of perception-aware planners in the literature plan  multicopter trajectories by solving an optimization problem,} encoding the perception-awareness as a cost term or constraint.
In \cite{featureVisible1}, the authors add the visibility of features as a constraint in the optimization of B-spline trajectories. The differential flatness property of multicopters is used to speed up the optimization. 
\mmod{The method is validated in simulation.}
In \cite{featureVisible3, featureVisible2}, the vehicle's trajectories are planned while maintaining a given set of landmarks within the field of view of its on-board camera. 
The first step is geometric path planning, followed by a time parameterization of the planned path to satisfy the kinodynamic constraints of the quadcopter.
In addition to the visibility of the features, maximizing the covisibility of features is also helpful in reducing \mmod{the} state estimation error \cite{featureCovisibility}.
The goal is to keep features visible in the camera field of view from one keyframe to the next, instead of to maximize the visibility of features in each image.
The authors first plan a minimum snap trajectory with only position, and then the yaw angle is planned to maximize the features' covisibility. 
\mmod{Indoor experiments are shown, with the locations of the landmarks in the environment are known a priori.}
Authors of \cite{PAMPC} additionally take the feature's movement speed into account by adding a perception cost term to reduce the movement speed of the feature points' centroid in the image and keep it close to the image center.
Model predictive control (MPC) is used for trajectory planning and the  optimization is accelerated by a sequential quadratic program (SQP) approximation.
In \cite{semantic}, semantic information is used to plan trajectories in areas with more texture and to avoid places with unreliable visual features, such as lakes.
\mmod{Simulation results are shown with the semantic information from the ground truth of the simulator.}

\par
\mmod{Another major category of perception-aware planners use sampling-based methods,} where the sampled trajectories are evaluated based on both the mission goal and their impact on state estimation. 
\mmod{In \cite{perc-aware-add1}, the authors use the rapidly exploring random belief tree (RRBT) approach to evaluate multiple candidate paths in a known map and select paths with minimum pose uncertainty.
The evaluation of the paths is done offline before the flight because of high computational cost.
The pose covariance of the vehicle is estimated via bundle adjustment, by minimizing the reprojection errors of the 3D map points across all images (augmented by noise with constant covariance).
The authors of \cite{perc-aware-add2} propose a planner that prefers feature-rich regions,by introducing a viewpoint score based on the visibility of the visual features and incorporating this score in RRT*.
}
Perception-awareness is considered in \cite{stateEstProp1, stateEstProp2} to  improve the mapping and state estimation accuracy during quadcopter exploration.
The outer layer planner generates paths that explore the space using the  rapidly-exploring random tree (RRT), and the inner layer planner aims to improve the mapping and state estimation accuracy.
The authors propagate the state estimation for different paths found by the inner layer planner, and choose the one which minimizes the state estimation uncertainty. 
In \cite{PnP}, the task of reaching a given goal with the highest accuracy while avoiding obstacles in the environment is investigated. 
The planner generates candidate trajectories and evaluates them in terms of perception quality, collision probability, and distance to the goal.
Given each sampled trajectory, the authors simulate the  observed features if the trajectory is followed and construct a least squares problem to estimate the vehicle's pose estimation error.

\par
\mmod{Unlike most of the work in the literature using feature-based methods for localization, the authors in \cite{perc-aware-add3} propose a perception-aware planner for direct visual inertial odometry.
The direct VIO method estimates the vehicle's pose based on the image pixels'  intensities and an introduction of it can be found at e.g. \cite{direct-method}.
They adapt the RRT* framework to select trajectories that minimize the camera pose uncertainty according to photometric information (computed with dense image-to-model alignment).
In the aforementioned papers, most use indoor experiments or simulations to validate their proposed methods, while both indoor and outdoor experiments are conducted in \cite{perc-aware-add1, stateEstProp2}.
}

\par
In this work, we focus on the problem of flying a multicopter to a desired position at a high speed, avoiding obstacles in the environment, while achieving good state estimation accuracy from the VIO.
A stereo depth camera with an IMU is used for collision avoidance and VIO.
To quickly check if a trajectory is collision-free, we use a sampling-based planner named RAPPIDS \cite{rappids1, rappids2}. 
In the next step, given a sampled trajectory that is collision-free, we predict the pose of the vehicle (assuming perfect trajectory tracking) and then the position and velocity of the VIO features in the camera frame.
The vehicle pose estimation is constructed as a least-squares problem, and each feature's variance is estimated from its velocity in the image.
We then evaluate the perception cost of the trajectory based on the vehicle's predicted position estimation uncertainty if that trajectory is followed. 
In addition, the speed cost of the trajectory is the negative value of its average speed towards the goal.
The trajectory that minimizes the perception cost plus the speed cost is selected as the trajectory to follow.
Our planner runs in a receding horizon manner, and it replans every time a new image arrives.
\par
Our proposed planner can reduce the state estimation error of the VIO by planning trajectories that guide the vehicle towards feature-rich areas and by preventing the vehicle from executing overly aggressive trajectories (causing  motion blur).
Meanwhile, the trajectories it plans are collision-free and dynamically feasible.
The planner is also computationally efficient enough to run on an onboard embedded computer in real-time.
Compared with the existing works in the literature, the contributions of this work are:
\begin{enumerate}
    \item We propose a \mmod{trajectory planner} generating collision-free trajectories that guides the vehicle towards the target \mmod{in unknown environments,} while avoiding regions with few visual features, \mmod{preventing overly aggressive flights, and avoiding obstacles.}
    \item We propose a perception cost function considering both the motion blur of the features and their locations.
    \item \mmod{Automatic} adaptation of the trajectory's aggressiveness under environments with different light levels.
    \item Experimental validation in both indoor and outdoor environments \mmod{with the algorithm running onboard in real-time, validating the effectiveness of the proposed method.}
\end{enumerate}

\section{System overview}
\label{sec:system_overview}
\begin{figure}[!tp]
    \centering
    \includegraphics[width =1.0\linewidth]{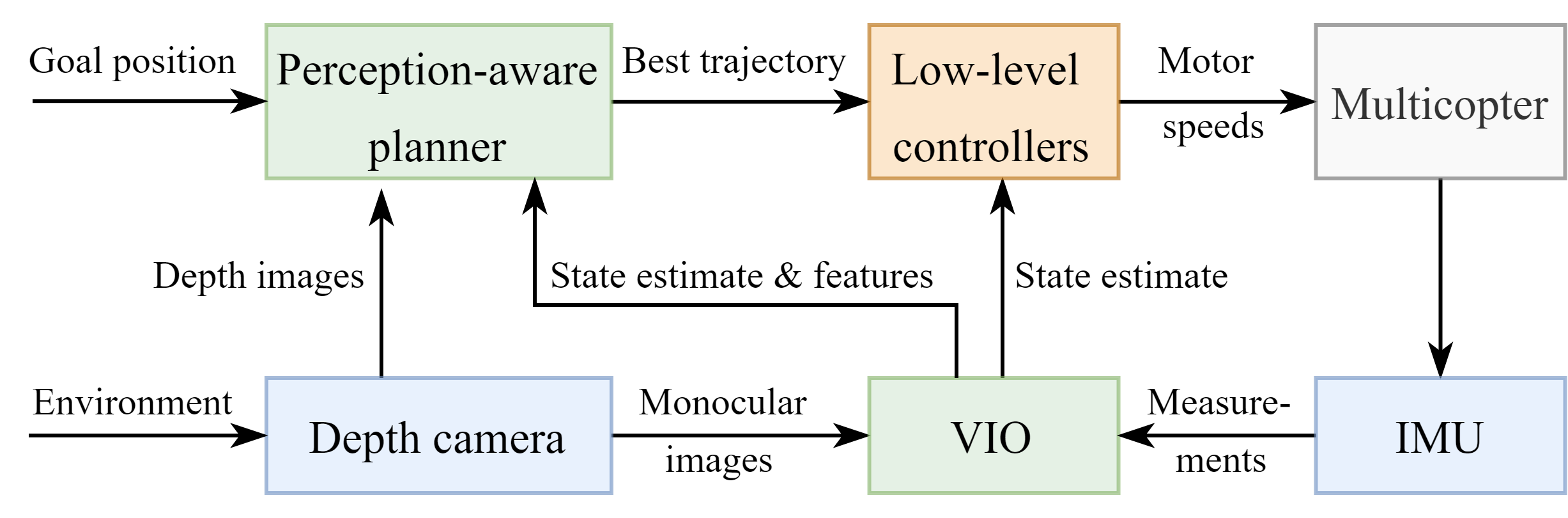}
    \caption{\mmod{Block diagram of the proposed perception-aware planning system for multicopters. 
    The perception-aware planner and VIO (with green background) run on the onboard computer.
    The goal position for the perception-aware planner could be given by the user, or by a high-level global planner.}}
    \label{fig:syste_overview}
\end{figure}

In this section, we give a brief overview of the proposed perception-aware planning system, a block diagram of which is shown in \figref{fig:syste_overview}.
The goal is to fly the drone to a goal position at a fast speed, avoid obstacles in the way, and achieve a good state estimation accuracy from the VIO.
\mmod{The goal position input to the perception-aware planner could be given directly by the user, or be given by a high-level global planner when used in an autonomous navigation framework.}
The environment is assumed to be stationary.
\par
The perception-aware planner uses the depth images from the stereo depth camera to detect obstacles in the environment.
We use the RAPPIDS planner \cite{rappids1, rappids2}, a memory-less sampling-based planner, to generate a group of collision-free candidate trajectories at low computational cost.
We use OpenVINS \cite{openvins} for VIO, which uses monocular images  from the depth camera and IMU measurements to estimate the state of the vehicle.
\mmod{It uses the monocular images from the left and right cameras on the stereo depth camera directly, instead of using the depth image for state estimation.
Our proposed planner does not depend on a specific VIO method, and other feature-based VIO algorithms can also be used with it.}
The VIO also sends the \mmod{3D positions of the tracked features in the world frame} to the perception-aware planner to evaluate the perception cost $c_\text{perc}$ of each candidate trajectory.
\par
\mmod{
One of the major contributions of this paper is the proposal of a novel perception-aware cost $c_\text{perc}$, whose derivation is detailed in \secref{sec:preception_cost}.
The derived cost term not only considers the number and position of the VIO feature points, but also their uncertainty from motion blur.
With this cost term, our proposed planner plans trajectories leading to reduced VIO state estimation uncertainty.
Such trajectories guide the vehicle to fly close to areas with visual features to keep more feature points in the camera's FOV,
and they are not overly aggressive to cause excessive motion blur of the visual features.
The camera's exposure time is also considered in the perception cost: the cost term will be larger in darker environments where a longer exposure time makes  images more prone to motion blur.
This makes the planner prefer less-aggressive candidate trajectories in darker environments to prevent poor VIO accuracy.
}

In addition, we add a flight speed related cost $c_\text{speed}$ (defined in \secref{sec:speed_cost}) to encourage fast flight towards the goal.
A parameter $k_\text{perc}$ is introduced to determine the weight of the perception quality, and the collision-free trajectory with the minimum total cost $k_\text{perc} c_\text{perc} + c_\text{speed}$ is chosen as the best trajectory.
\mmod{
The best trajectory is then tracked by the low-level position and attitude controllers of the multicopter.
}
\par
The perception-aware planner runs in a receding horizon manner, and it replans every time a new depth image is received.
The tracked trajectory is updated if a lower-cost trajectory is found.
\mmod{Recursive feasibility is guaranteed: all sampled trajectories have zero speed and acceleration at the end, and the planner assumes no other visual features except those already found by the VIO.
As a result, in the worst case, the vehicle will follow the current trajectory to the end and stop safely if no new trajectories are found, avoiding collision or flying into areas without features.
}

\section{Perception-aware cost derivation}
\label{sec:preception_cost}

\begin{figure}[!htp]
    \centering
    \includegraphics[width = \linewidth]{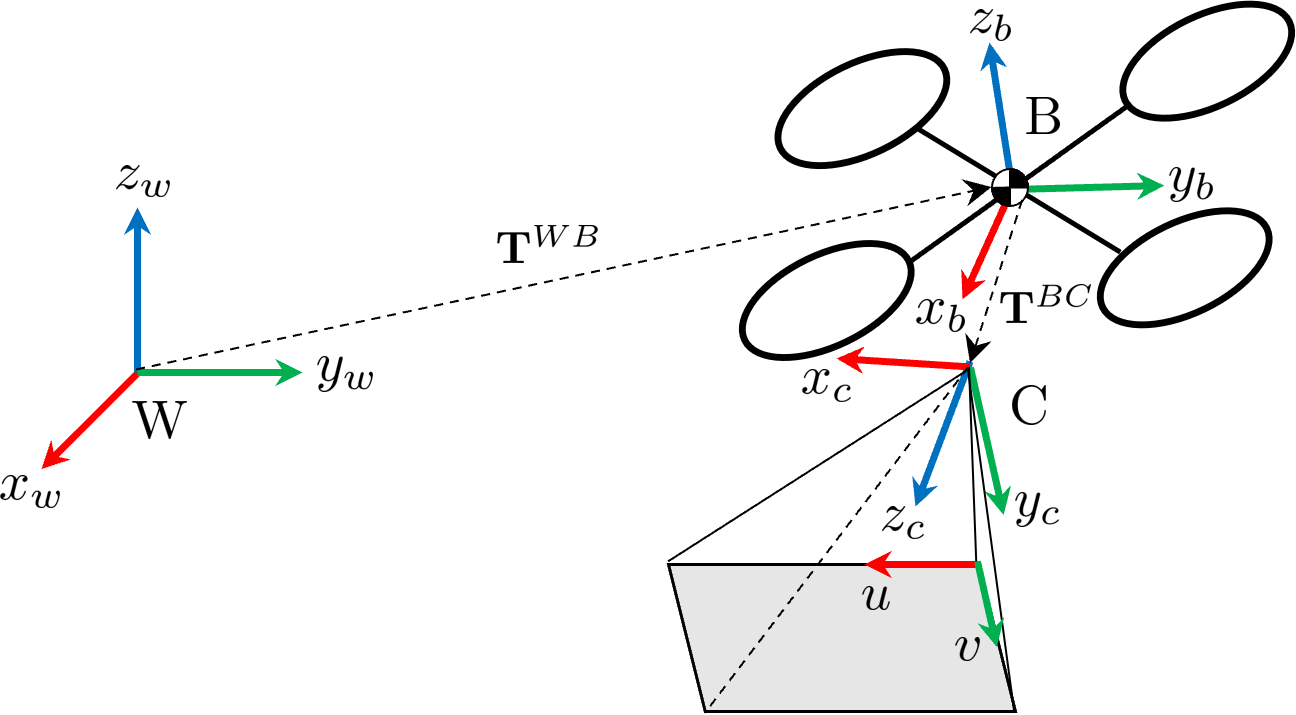}
    \caption{An illustration of the world frame W, body frame B and camera frame C. The position and attitude of the vehicle's body frame with respect to the world frame is given by $\bm{T}^{WB}$. The position and attitude of the camera frame in the body frame is given by $\bm{T}^{BC}$, which is a rigid body transformation.}
    \label{fig:frames}
\end{figure}

In this section, we introduce the derivation of the perception cost \mmod{that helps to reduce the state estimation uncertainty.}
We define the world frame W, body frame B, and camera frame C, as shown in \figref{fig:frames}.
Vector and matrix variables are written in boldface.
The notations used in this section are summarized in Table \ref{tab:symbols}.
\begin{table}[!htp]
	\def\arraystretch{1.2}
	\caption{Notations used in \secref{sec:preception_cost}.}
	\vspace{-1ex}
	\label{tab:symbols}
	\centering
\begin{tabular}{|l|l|}
\hline
\multicolumn{1}{|c|}{Symbol} & \multicolumn{1}{c|}{Meaning} \\ \hline
$c_\text{perc}, c_\text{speed}, c_\text{tot}$ & perception, speed, and total cost \\ \hline
$k_\text{perc}$ & weight coefficient of perception cost \\ \hline
$W, B, C$ & \begin{tabular}[c]{@{}l@{}}world frame, vehicle body frame, camera\\ frame\end{tabular} \\ \hline
$\bm{T}^{WB} = [\bm{R}^{WB} \;  \bm{t}^{WB}]$ & \begin{tabular}[c]{@{}l@{}}pose of the vehicle body frame B \\ in the world frame W\end{tabular} \\ \hline
$\bm{T}^{BC} = [\bm{R}^{BC} \; \bm{t}^{BC}]$ & \begin{tabular}[c]{@{}l@{}}pose of the camera frame C in vehicle \\ body frame B (rigid transformation)\end{tabular} \\ \hline
$\bm{T}^{CW} =[\bm{R}^{CW} \;  \bm{t}^{CW}]$ & \begin{tabular}[c]{@{}l@{}}camera extrinsics: pose of the world\\ frame W in the camera frame C\end{tabular} \\ \hline
subscript $j$ & the $j$th sampled pose of the trajectory \\ \hline
subscript $k$ & the $k$th VIO feature \\ \hline
$\bm{I}_j$ & set of visible features at sampled pose $j$ \\ \hline
\begin{tabular}[c]{@{}l@{}}$\bm{\Gamma}(t)$ \\ $= [x(t) \; y(t) \;  z(t) \;  \psi(t)]^T$\end{tabular} & \begin{tabular}[c]{@{}l@{}}a trajectory consisted of 3D positions \\ $x(t), y(t), z(t)$, and yaw angle $\psi(t)$\end{tabular} \\ \hline
\begin{tabular}[c]{@{}l@{}}$\bm{P}_{k} = [X_{k} \; Y_{k} \; Z_{k}]^T$, \\ $\bm{P}_{kj}^{'} = [X^{'}_{kj} \; Y^{'}_{kj} \; Z^{'}_{kj}]^T$\end{tabular} & \begin{tabular}[c]{@{}l@{}}3D coordinates of the $k$th feature in W,\\ 3D coordinates of the $k$th feature in  C\\ when vehicle is at sampled pose $j$\end{tabular} \\ \hline
\begin{tabular}[c]{@{}l@{}}$\hat{\bm{b}}_{kj} = [\hat{u}_{kj} \; \hat{v}_{kj}]^T$, \\ $ \bm{b}_{kj} = [u_{kj} \; v_{kj}]^T$\end{tabular} & \begin{tabular}[c]{@{}l@{}}$k$th feature's observation in image, \\ estimated projection in the image  \\ (when vehicle is at sampled pose $j$)\end{tabular} \\ \hline
$S(\cdot)$ & \begin{tabular}[c]{@{}l@{}}function converting a $\mathbb{R}^3$ vector to its \\ corresponding skew symmetric matrix\end{tabular} \\ \hline
$\bm{\xi}$ & error of estimated camera extrinsics \\ \hline
$\bm{J}_{\bm{\xi},kj}$ & \begin{tabular}[c]{@{}l@{}}${\partial{\bm{b}}_{kj}} / {\partial{\bm{\xi}}}$, Jacobian of  the $k$th feature\\ with respect to $\bm{\xi}$ at sampled pose $j$\end{tabular} \\ \hline
$\bm{\Sigma}_{\bm{\xi}j}$ & \begin{tabular}[c]{@{}l@{}}covariance of estimated $\bm{\xi}$ \\ at sampled pose $j$\end{tabular} \\ \hline
$\bm{v}_j,  \bm{\omega}_j$ & \begin{tabular}[c]{@{}l@{}}vehicle's velocity, angular velocity \\ at sampled pose $j$\end{tabular} \\ \hline
$\sigma_{kj,\parallel}^2,  \sigma_n^2$ & \begin{tabular}[c]{@{}l@{}}feature's variance because of speed,\\ feature's variance because of vibration\end{tabular} \\ \hline
$ \bar{\dot{\bm{b}}}_{kj} = [\bar{\dot{u}}_{kj} \; \bar{\dot{v}}_{kj}]^T$ & \begin{tabular}[c]{@{}l@{}}the normalized speed of the $k$th feature\\ in image at sampled pose $j$\end{tabular} \\ \hline
$t_\text{exp}$ & camera's exposure time \\ \hline
$\bm{\Sigma}_{\bm{b}, kj}$ & \begin{tabular}[c]{@{}l@{}}covariance of the $k$th feature in the image\\ at sampled pose $j$\end{tabular} \\ \hline
$\bm{\Sigma}_{\bm{b}j}$ & \begin{tabular}[c]{@{}l@{}}covariance of visible features in the image\\ at sampled pose $j$\end{tabular} \\ \hline
\end{tabular}
\end{table}
\par

The pose of the vehicle's body frame B with respect to the world frame W is given by $\bm{T}^{WB} = \mmat{\bm{R}^{WB} & \bm{t}^{WB}}$, where $\bm{R}^{WB}$ and $\bm{t}^{WB}$ are the orientation and position of B with respect to W.
We assume that the vehicle is equipped with a depth camera, whose pose in the vehicle's body frame is given by a rigid body transformation $\bm{T}^{BC} = \mmat{\bm{R}^{BC} & \bm{t}^{BC}}$. 
The extrinsics of the camera is given by $\bm{T}^{CW} = \mmat{\bm{R}^{CW} & \bm{t}^{CW}}$.
\par

The perception cost of a trajectory $c_\text{perc}$ represents  the uncertainty of vehicle position estimation if the trajectory is followed. 
Similarly to \cite{PnP}, we estimate the variance of vehicle pose estimation by formulating a least-squares problem.
However, we also estimate the variance of the feature points in the image according to their movement speed in the image, instead of assuming a constant variance for the features as in \cite{PnP}.
This helps to take motion blur into account and discourages the multicopter from executing an overly aggressive trajectory, which would adversely affect the VIO accuracy.
The function mapping the pose estimation variance to the perception cost $c_\text{perc}$ is also different, for faster computation and more intuitive tuning of the perception cost's weight coefficient $k_\text{perc}$.

\subsection{Cost of a trajectory}
\begin{figure}[b]
    \begin{center}
    \includegraphics[width=0.95\columnwidth]{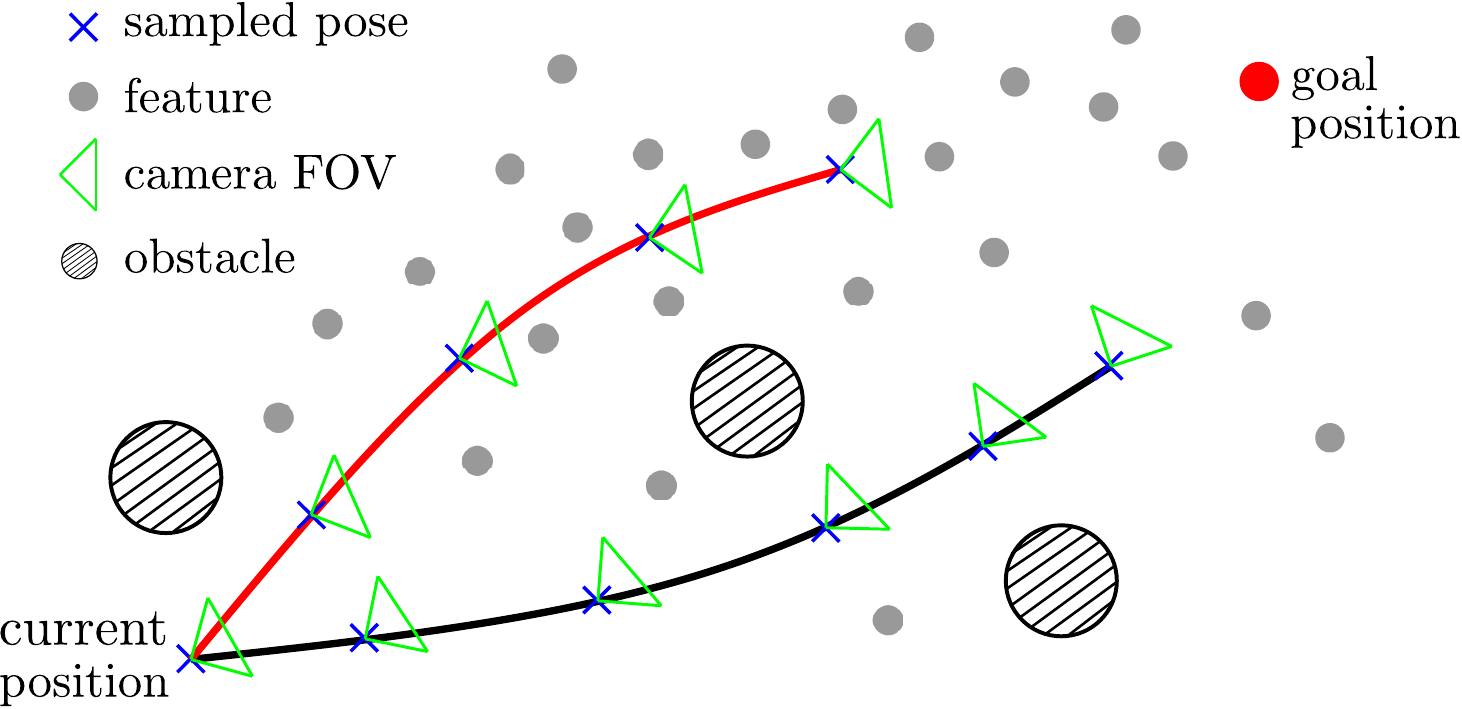}
    \end{center}
    \caption{We discretize sampled trajectories at a fixed time interval to evaluate their perception cost. 
    When the camera's exposure time is short and motion blur is not significant, the trajectory in red has lower perception cost than the trajectory in black. 
    Because it keeps more features in the camera's view and is also closer to the features.
    \mmod{On the trajectories, the camera always points towards the target point to better see the obstacles.}}
    \label{fig:sampled_poses}
\end{figure}

Given a trajectory $\bm{\Gamma}(t) = \mmat{x(t) & y(t) & z(t) & \psi(t)}^T$ and the 3D positions of VIO features $F_\text{feature} := \{\bm{P}_k\}^M_{k=1}$ in the world frame, we need to predict the pose estimation error if the given trajectory is followed.
In $\bm{\Gamma}(t)$, the 3D positions of the vehicle are represented as $\mmat{x(t) & y(t) & z(t)}^T$, and the yaw angle of the vehicle is represented as $\psi(t)$.
We first sample the poses of the vehicle $\{\bm{T}^{WB}_j\}_{j=1}^N$ along the trajectory at a fixed time interval.
For a sampled pose $\bm{T}^{WB}_j$, we can find the extrinsics of the camera $\bm{T}^{CW}_j$ and thus the visible features in $F_\text{feature}$ , as shown in \figref{fig:sampled_poses}.
We denote the indices of visible features at the camera pose  $\bm{T}^{CW}_j$ as $\bm{I}_j$.
\par
Then if the vehicle moves to $\bm{T}^{WB}_j$, the extrinsics of the camera $\bm{T}^{CW}_j$ can be estimated by the following least squares problem:
\begin{equation}
    \bm{T}^*_j = \text{arg} \min_{\bm{T}} \sum_{k \in \bm{I}_j}{\mnorm{\hat{\bm{b}}_{kj} - \text{proj}\left( \bm{T} \bm{P}_k \right)}_2^2}, \label{eq:ls1} 
\end{equation}
where $\text{proj}(\cdot)$ represents the projection function of the camera, and $\hat{\bm{b}}_{kj} = \mmat{\hat{u}_{kj} & \hat{v}_{kj}}^T$ is the $k$th feature point's observation in the image, which has noise due to motion blur and camera lens imperfections.
\par
The optimization problem \eqref{eq:ls1} can usually be solved in an iterative way. 
It can be converted into the following form:
\begin{equation}
    \bm{\xi}^* = \text{arg} \min_{\bm{\xi}} \sum_{k \in \bm{I}_j}{\mnorm{\hat{\bm{b}}_{kj} - \text{proj}\left( \text{exp}\left(\bm{\xi}\right) \bm{T}^{CW}_j  \bm{P}_k \right)}_2^2},  \label{eq:ls2}
\end{equation}
where $\bm{\xi}$ is in $\mathfrak{se}(3)$ and represents the error of the estimated camera extrinsics.
The function $\text{exp}(\cdot)$ maps $\mathfrak{se}(3)$ to $SE(3)$.
The estimated camera extrinsics $\bm{T}^{CW}_j$ is updated iteratively by $\bm{T}^{CW}_j = \text{exp}\left( \bm{\xi} \right) \bm{T}^{CW}_j$.
\par
Define $\bm{b}_{kj} = \text{proj}\left( \bm{T}^{CW}_j \bm{P}_k \right)$, which is the projected coordinates of the $k$th feature in the image frame.
The equation \eqref{eq:ls2} can be solved through linearization at the current estimation of $\bm{T}^{CW}_j$:
\begin{equation}
\begin{split}
    \bm{\xi}^* &= \text{arg} \min_{\bm{\xi}} \sum_{k \in \bm{I}_j} \mnorm{ \hat{\bm{b}}_{kj} - \text{proj}\left( \bm{T}^{CW}_j \bm{P}_k \right) - \bm{J}_{\bm{\xi},kj} \bm{\xi}}_2^2 \\ 
    &= \text{arg} \min_{\bm{\xi}} \sum_{k \in \bm{I}_j} \mnorm{ \hat{\bm{b}}_{kj} - \bm{b}_{kj} -  \bm{J}_{\bm{\xi},kj} \bm{\xi}}_2^2,  \label{eq:ls3} 
\end{split}
\end{equation}
where $\bm{J}_{\bm{\xi},kj} = \frac{\partial{\bm{b}}_{kj}}{\partial{\bm{\xi}}}, \; k \in \bm{I}_j$.
\par
Denote $\bm{P}_{kj}^{'} = \mmat{X_{kj}^{'} & Y_{kj}^{'} & Z_{kj}^{'}}^T = \bm{T}^{CW}_j \bm{P}_k$ as the coordinates of features in the camera frame, then we have:
\begin{equation}
    \bm{J}_{\bm{\xi},kj} = \frac{\partial{\bm{b}}_{kj}}{\partial{\bm{\xi}}} = \frac{\partial \bm{b}_{kj}}{\partial \bm{P}_{kj}^{'}} \frac{\partial \bm{P}_{kj}^{'}}{\partial \bm{\xi}}. \label{eq:jacobian}
\end{equation}
Denote the camera's focal length as $f_x, f_y$ and its focal point's coordinates in the image as $\mmat{c_x & c_y}^T$, from the pinhole camera model, we have:
\begin{equation}
    \bm{b}_{kj} = \mmat{u_{kj} \\ v_{kj}} = \text{proj}\left( \bm{P}_{kj}^{'} \right) 
    = \mmat{f_x \frac{X_{kj}^{'}}{Z_{kj}^{'}} + c_x \\
    f_y \frac{Y_{kj}^{'}}{Z_{kj}^{'}} + c_y }. \label{eq:pinhole_model}
\end{equation}
Besides, with some Lie algebra derivations, we can get
\begin{equation}
    \frac{\partial \bm{P}_{kj}^{'}}{\partial \bm{\xi}} = \mmat{\bm{I}_{3\times3} &     -S \left( \bm{P}_{kj}^{'} \right) }, \label{eq:feature_lie_der}
\end{equation}
where the function $S \left( \cdot \right)$ converts a vector in $\mathbb{R}^3$ to its corresponding 3-by-3 skew-symmetric matrix.
By combining \eqref{eq:jacobian}, \eqref{eq:pinhole_model} and \eqref{eq:feature_lie_der}, we can get the expression of the Jacobian
\begin{equation}
\setlength\arraycolsep{3pt}
\begin{split}
    & \bm{J}_{\bm{\xi},kj} = \\
    & \mmat{\frac{f_x}{Z^{'}_{kj}} & 0 & -\frac{f_x X^{'}_{kj}}{{Z^{'}_{kj}}^2} & -\frac{f_x X^{'}_{kj} Y^{'}_{kj}}{{Z^{'}_{kj}}^2} & f_x + \frac{f_x {X^{'}_{kj}}^2}{{Z^{'}_{kj}}^2} &  -\frac{f_x Y^{'}_{kj}}{Z^{'}_{kj}} \\
    0 & \frac{f_y}{Z^{'}_{kj}} & -\frac{f_y Y^{'}_{kj}}{{Z^{'}_{kj}}^2} & -f_y - \frac{f_y {Y^{'}_{kj}}^2}{{Z^{'}_{kj}}^2} & \frac{f_y X^{'}_{kj} Y^{'}_{kj}}{{Z^{'}_{kj}}^2} & \frac{f_y X^{'}_{kj}}{Z^{'}_{kj}} } \label{eq:jacobian_expanded}
\end{split}
\end{equation}
\par
As a result, the covariance of the estimated parameter $\bm{\xi}$ in \eqref{eq:ls3} is given by 
\begin{equation}
    \bm{\Sigma}_{\bm{\xi}j} = (\bm{J}_{j}^T \bm{J}_j)^{-1} \bm{J}_{j}^T \bm{\Sigma}_{\bm{b}j} \bm{J}_j (\bm{J}_j^T \bm{J}_j)^{-1} = (\bm{J}_j^T \bm{\Sigma}_{\bm{b}j}^{-1} \bm{J}_{j})^{-1}, \label{eq:pos_var}
\end{equation}
where the matrix $\bm{J}_{j}$ is stacked up of $\bm{J}_{\bm{\xi},kj}$, and $\bm{\Sigma}_{\bm{b}j}$ is the covariance of visible features, which is related to the feature's speed in the image plane and is derived in the following \secref{sec:feature_var}.
\par
For computational efficiency and more intuitive tuning of $k_\text{perc}$, we only consider the estimation variance of the first three elements of $\bm{\xi}$, which corresponds to the estimated position of the vehicle. 
The perception cost of a trajectory is defined as:
\begin{equation}
    c_\text{perc} = \sum_{j=1}^{N} \frac{\sqrt{\bm{\Sigma}_j^{(1,1)}} + \sqrt{\bm{\Sigma}_j^{(2,2)}} + \sqrt{\bm{\Sigma}_j^{(3,3)}}}{3N}, \label{eq:perc_cost_def}
\end{equation}
which corresponds to the mean sum of per-axis standard deviation of \mmod{the position estimates} over the sampled times.

\subsection{Feature variance estimation}
\label{sec:feature_var}
When the vehicle moves to $\bm{T}^{WB}_j$, the 3D position of a visible VIO feature in the camera's frame $\bm{P}_{kj}^{'} = \mmat{X_{kj}^{'} & Y_{kj}^{'} & Z_{kj}^{'}}^T$ is given by: 
\begin{equation}
    \bm{P}_{kj}^{'} = \bm{T}^{CW}_j \bm{P}_k = \bm{R}^{CW}_j \left( \bm{P}_k - \bm{R}^{WB}_j \bm{t}^{BC} - \bm{t}^{WB}_j \right). \label{eq:feature_cam_pos}
\end{equation}
To obtain the feature's velocity in the camera frame, we differentiate \eqref{eq:feature_cam_pos} with respect to time:
\begin{equation}
\begin{split}
    \dot{\bm{P}}_{kj}^{'}
    = & \mmat{\dot{X}_{kj}^{'} & \dot{Y}_{kj}^{'} & \dot{Z}_{kj}^{'}}^T \\
    = & - \bm{R}^{CB} S(\bm{\omega}_j) \bm{R}^{CB} \bm{P}_k^{'} \\
      & - \bm{R}^{CB} S(\bm{\omega}_j) \bm{t}^{BC} - \bm{R}_j^{CW} \bm{v}_j^{WB}
\end{split}
\end{equation}
where $\bm{v}_j^{WB}$ is the vehicle's velocity in the world frame and $\bm{w}_j$ is the vehicle's angular velocity in the body frame.
They can be predicted given a trajectory $\bm{\Gamma}(t)$ \cite{min_snap}.
Also, $\bm{R}^{CB} = \left(\bm{R}^{BC} \right)^{-1}$.
\par
We then differentiate \eqref{eq:pinhole_model} with respect to time, to get the feature's speed in the image plane:
\begin{equation}
\begin{split}
    \dot{u}_{kj} = f_x \frac{\dot{X}_{kj}^{'} Z_{kj}^{'} - X_{kj}^{'} \dot{Z}_{kj}^{'}}{{Z_{kj}^{'}}^2}, \;
    \dot{v}_{kj} = f_y \frac{\dot{Y}_{kj}^{'} Z_{kj}^{'} - Y_{kj}^{'} \dot{Z}_{kj}^{'}}{{Z_{kj}^{'}}^2}. \label{eq:feature_speed}
\end{split}
\end{equation}

\par
Due to the movement of the camera, a feature point will have some motion blur in the image.
Denote the camera's exposure time as $t_\text{exp}$, and the feature could be approximated by a straight line of length $t_\text{exp} \: \sqrt{ \dot{u}^2_{kj} + \dot{v}^2_{kj}}$ when the exposure time is relatively short. 
We assume that the feature point is uniformly distributed on this straight line, whose direction is the feature's speed ($\dot{\bm{b}}_{kj}$) direction in the image plane.
The feature's variance in this direction is approximated by:
\begin{equation}
    \sigma_{kj,\parallel}^2 = \frac{t_\text{exp}^2 \left( \dot{u}^2_{kj} + \dot{v}^2_{kj} \right)}{12}.
\end{equation}
In addition, due to vehicle vibration and the imperfect lens, the feature has an additional variance $\sigma_{n}^2$, which we assume to be omnidirectional and can be measured experimentally.
\begin{figure}[!htp]
    \begin{center}
    \includegraphics[width=\columnwidth]{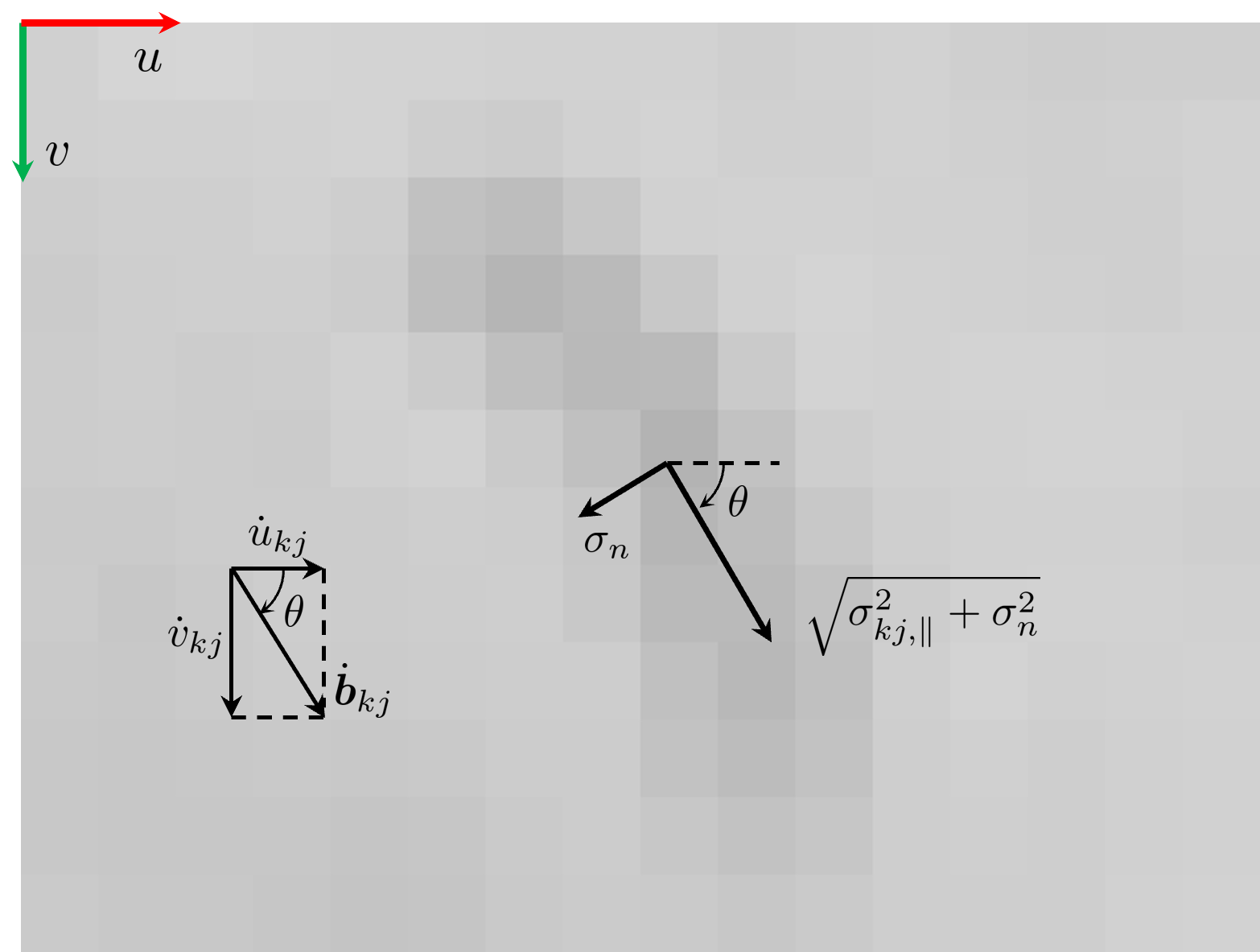}
    \end{center}
    \caption{A feature point becomes a blurry line in the image because of motion blur. The angle of the line is $\theta$, which is the same as the feature's speed in the image plane $\dot{\bm{b}}_{kj}$. The standard deviation of the feature's position on the speed direction is $\sqrt{\sigma_{kj,\parallel}^2 + \sigma_n^2}$, and is $\sigma_n$ on the perpendicular direction.}
    \label{fig:blurred_feature}
\end{figure}
\par
Define the normalized feature speed in the image to be $ \bar{\dot{\bm{b}}}_{kj} = \mmat{\bar{\dot{u}}_{kj} & \bar{\dot{v}}_{kj}}^T$.
Then, the covariance of the feature in the image plane is given by:
\begin{equation}
\setlength\arraycolsep{3pt}
\begin{split}
    & \bm{\Sigma}_{\bm{b}, kj} = \mmat{\bar{\dot{u}}_{kj} & -\bar{\dot{v}}_{kj} \\ \bar{\dot{v}}_{kj} & \bar{\dot{u}}_{kj}} \mmat{\sigma_{kj,\parallel}^2 + \sigma_{n}^2 & 0 \\ 0 & \sigma_{n}^2} \mmat{\bar{\dot{u}}_{kj} & -\bar{\dot{v}}_{kj} \\ \bar{\dot{v}}_{kj} & \bar{\dot{u}}_{kj}}^T = \\
    & \mmat{\bar{\dot{u}}_{kj}^2 \sigma_{kj,\parallel}^2 + \left( \bar{\dot{u}}_{kj}^2 + \bar{\dot{v}}_{kj}^2 \right) \sigma_{n}^2 & \bar{\dot{u}}_{kj} \bar{\dot{v}}_{kj} \sigma_{kj,\parallel}^2 \\
    \bar{\dot{u}}_{kj} \bar{\dot{v}}_{kj} \sigma_{kj,\parallel}^2 & \bar{\dot{v}}_{kj}^2 \sigma_{kj,\parallel}^2 + \left( \bar{\dot{u}}_{kj}^2 + \bar{\dot{v}}_{kj}^2 \right) \sigma_{n}^2}  \label{eq:feature_cov} 
\end{split}
\end{equation}
We can then get the covariance matrix of the visible features in \eqref{eq:pos_var}:
\begin{equation}
    \bm{\Sigma}_{\bm{b}j} = \mmat{\bm{\Sigma}_{\bm{b}, k_1j} & 0 & \dots & 0 \\ 
                                  0 & \bm{\Sigma}_{\bm{b}, k_2j} & \dots & 0 \\
                                  \vdots & \vdots & \ddots & \vdots \\ 
                                  0 & 0 & \dots & \bm{\Sigma}_{\bm{b}, k_mj}}, \label{eq:feature_var}
\end{equation}
where $k_1, k_2, \dots, k_m \in \bm{I}_j$.
By substituting \eqref{eq:feature_var} into \eqref{eq:pos_var} and \eqref{eq:perc_cost_def}, we can get the perception cost $c_\text{perc}$ of a sampled trajectory.

\section{Perception-aware trajectory planning}
In this section, we briefly introduce the RAPPIDS planner that we use to generate collision-free trajectories, which is first introduced in \cite{rappids1} and is improved in \cite{rappids2}. In addition, this section introduces how the perception-aware cost is integrated with the RAPPIDS planner to reduce the multicopter's state estimation uncertainty of VIO.

\subsection{RAPPIDS planner overview}
\subsubsection{Trajectory sampling}
\label{sec:traj-sampling}
The RAPPIDS planner samples fifth-order minimum jerk polynomial trajectories $\bm{s}(t)$ with different duration $T$ and end position $\bm{s}_{T}$ using a computationally efficient planner proposed in \cite{min_jerk}.
The sampled trajectories then go through checks to find if they are input feasible, below the flight speed limit (for safety), and collision-free.
The planner replans when a new depth image arrives, to take into account the latest obstacle information.
\par
The sampled trajectories are described as:
\begin{equation} \label{eqn:polynomial_trajectory}
    \bm{s}(t)=\frac{\bm{\alpha}}{120}t^5+\frac{\bm{\beta}}{24}t^4+\frac{\bm{\gamma}}{6}t^3+\frac{\ddot{\bm{s}}(0)}{2}t^2+\dot{\bm{s}}(0)t+\bm{s}(0),\ t\in[0, \; T]
\end{equation}
where $\bm{s}(0)$, $\dot{\bm{s}}(0)$, and $\ddot{\bm{s}}(0)$ are the position, velocity, and acceleration of the vehicle at the start time of the trajectory.
The terminal condition is selected to be at rest to ensure safety, which requires $\bm{s}(T) = \bm{s}_{T}$ and $\dot{\bm{s}}(T) = \ddot{\bm{s}}(T) = 0$.
Even if the planner cannot find a new collision-free trajectory (when new depth images arrive), the vehicle can continue following the current trajectory and stop safely.
The coefficients $\bm{\alpha}$, $\bm{\beta}$ and $\bm{\gamma}$ can be solved in closed form \cite{min_jerk}. 
\par
\mmod{
The yaw angle $\psi(t)$ of the trajectory is selected such that the multicopter always faces to the target point, to better detect the obstacles and free space on the way and thus increase the chance of finding trajectories with lower cost:
}
\begin{equation}
\begin{split}
    \psi(t) & = \text{arctan2} \left(\mmat{0 \; 1 \; 0} (\bm{s}_G - \bm{s}(t)), \; \mmat{1 \; 0 \; 0} (\bm{s}_G - \bm{s}(t)) \right), 
\end{split}
\end{equation}
where $\bm{s}_G$ and $\bm{s}(t)$ are the positions of the goal and the vehicle's position in the world frame, respectively.
\mmod{Note that we do not require the vehicle to be placed facing towards the goal point at the beginning of the flight, since the planner will adjust the vehicle's yaw towards the goal automatically.}

\subsubsection{Collision check}
\label{sec:collision-check}
\begin{figure}[b]
    \begin{center}
    \includegraphics[width=\columnwidth]{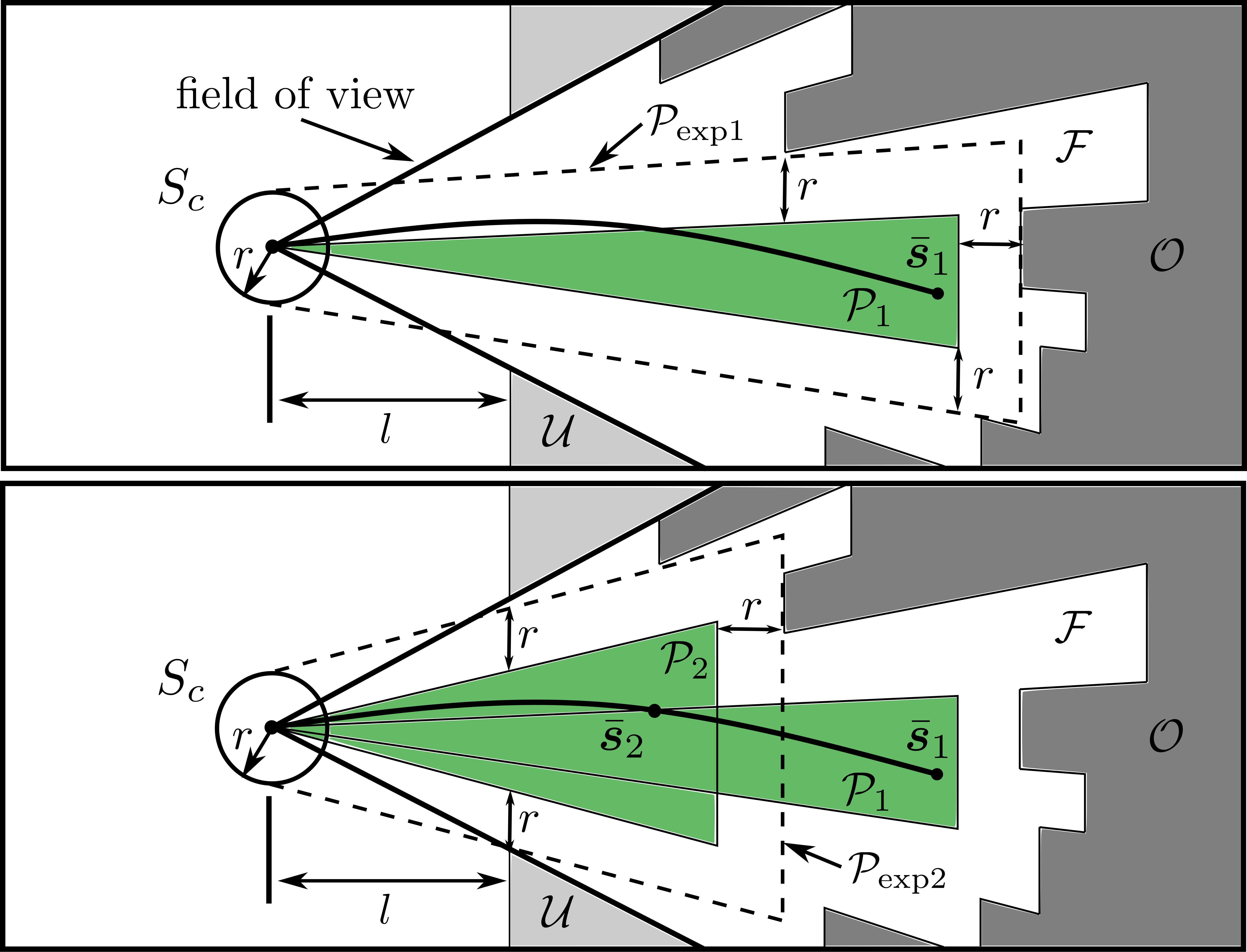}
    \end{center}
    \caption{
    The RAPPIDS planner partitions the free space with rectangular pyramids (shown in green) for fast collision check of a sampled trajectory.}
    \label{fig:depth_image}
\end{figure}

The RAPPIDS planner checks whether a sampled trajectory collides with obstacles by partitioning the free space into rectangular pyramids and checks if the trajectory remains in the union of these pyramids.
\mmod{
The idea of free space partitioning helps to reduce the computational cost of finding collision-free trajectories \cite{free-space-partition1, free-space-partition2, free-space-partition3, free-space-partition4}.}
In collision check, the vehicle is simplified as the smallest sphere $S_c$ containing the vehicle. 
The space partitioning process is illustrated in \figref{fig:depth_image}.
\label{sssec:collision-free}
Firstly, we can get the free space $\mathcal{F}$ and the occupied space $\mathcal{O}$ based on the depth image.
To avoid collisions with potential unseen obstacles, we treat all spaces $\mathcal{U}$ outside the depth camera's field of view that are $l$ distance away from the vehicle as occupied.
The next step is to search for the depth pixel closest to the end position of the trajectory $\bm{s}(T)$, marked as $\bar{\bm{s}}_1$ in the figure.
Then, starting with the nearest depth pixel and reading the surrounding depth pixels in a spiral sequence, we find the largest possible rectangular space $\mathcal{P}_{exp1}$ that does not intrude into the occupied space $\mathcal{O}$.
Finally, a pyramid $\mathcal{P}_1$ is created by shrinking the expanded pyramid $\mathcal{P}_{exp1}$ with the vehicle's radius $r$.
\par
Whether the sampled trajectory is within the union of generated collision-free pyramids can be determined efficiently using the method proposed in \cite{collision-check}.
If a sampled trajectory intersects with the union of existing pyramids, the algorithm tries to generate a new pyramid $\mathcal{P}_2$, starting the search from the intersection point, marked as $\bar{\bm{s}}_2$ in \figref{fig:depth_image}.
The pyramid generation process continues until the trajectory is within the union of the pyramids -- the trajectory is collision-free, or when no new pyramid could be generated -- the trajectory can collide with obstacles.
A detailed description of this algorithm can be found at \cite{rappids1}.

\subsection{Selection of the best trajectory}
\label{sec:speed_cost}
To encourage fast flight towards the target point (following \cite{rappids2}), we define the speed cost $c_\text{speed}$ as:
\begin{equation}
    c_\text{speed} = -\frac{\mnorm{\bm{s}_G-\bm{s}(0)}_2-\mnorm{\bm{s}_G-\bm{s}(T)}_2}{T},
\end{equation}
which is the negative of the average flight speed towards the goal for a sampled trajectory.
\par
There is a trade-off between fast flight and state estimation quality, especially in indoor environments where the camera's exposure time is long.
Flying at a fast speed will increase the motion blur, which makes the VIO less accurate and causes the perception cost $c_\text{perc}$ to increase.
As a result, we introduce a coefficient $k_\text{perc}$ to determine the importance of the VIO quality.
The larger we set $k_\text{perc}$, the more weight we put on the state estimation quality over fast flight and vice versa.
\par
\mmod{We choose the reference trajectory from the sampled trajectories that pass the RAPPIDS collision check (\secref{sec:collision-check}) to avoid obstacles.
Among them,} the trajectory with the minimum total cost $c_\text{tot} = k_\text{perc} c_\text{perc} + c_\text{speed}$ is chosen as the trajectory to follow.
The proposed perception-aware planner \mmod{runs} in a receding horizon manner, to take into account the latest information of the obstacles in the environment and the new feature points found by the VIO.
It tries to find a trajectory with lower total cost $c_\text{tot}$ every time a new depth image arrives and the current trajectory will continue to be followed if no better trajectory is found.
Since each generated trajectory has zero speed and acceleration at the end, and the planner assumes the only features in the environment are those found by the VIO, the planned trajectory is always safe.
The vehicle would stop safely if no new collision-free trajectory that allows the camera to see tracked VIO features could be found, avoiding the vehicle flying to areas with no VIO features.

\section{Experimental evaluation}
We validate the effectiveness of the proposed perception-aware planning method in both \mmod{stationary} indoor and outdoor environments, by comparing it with the original perception-agnostic RAPPIDS planner. 
The perception-aware planner is shown to improve the VIO's state estimation accuracy and the number of tracked features in the camera's field of view.
The experiment video can be found at \url{https://youtu.be/qO3LZIrpwtQ}.

\subsection{Hardware setup}
\begin{figure}[!htp]
    \begin{center}
    \includegraphics[width=0.9\columnwidth]{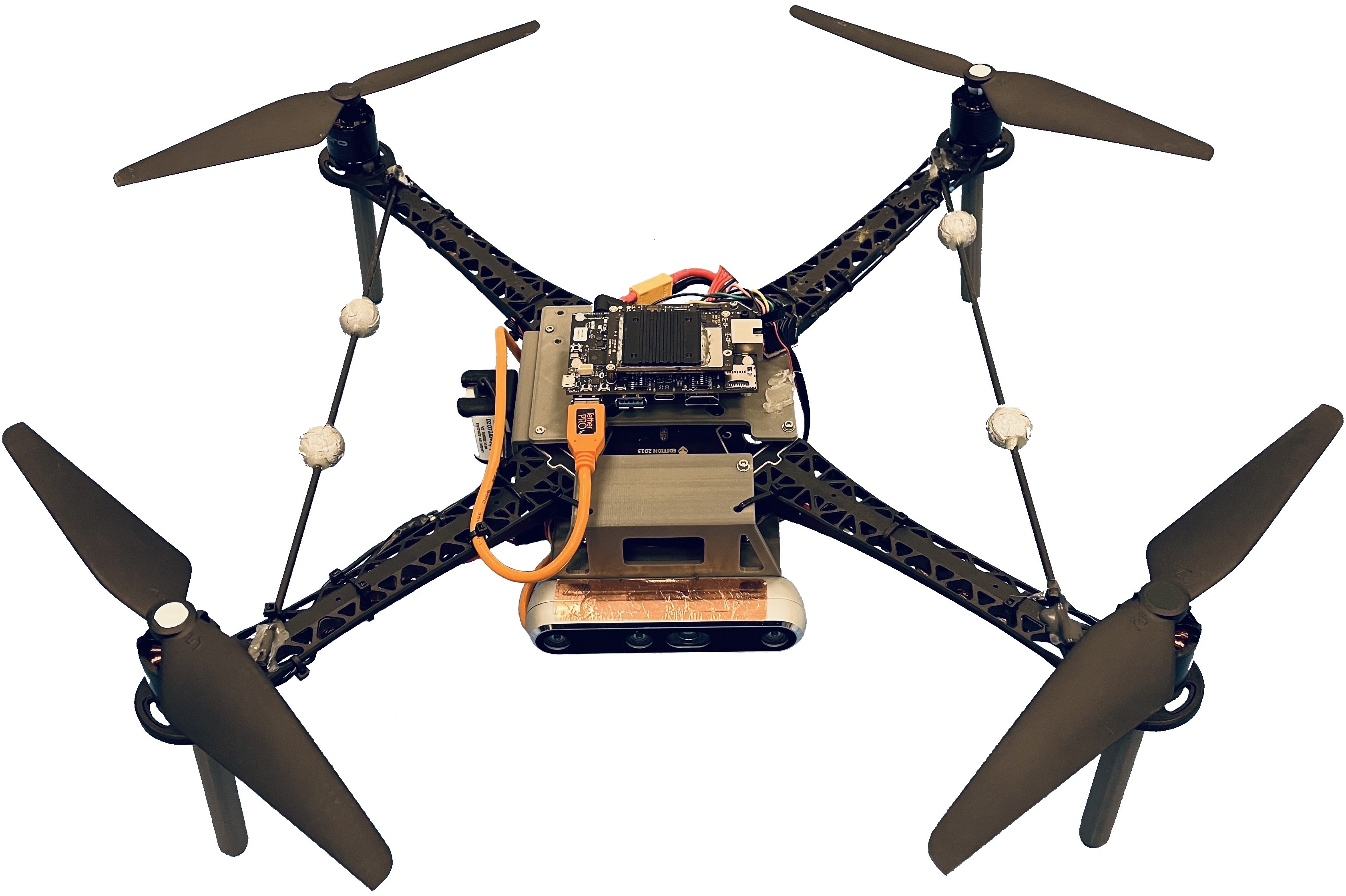}
    \end{center}
    \caption{The quadcopter used in the experiments. The distance between two diagonal motors is 500 mm.}
    \label{fig:drone_image}
\end{figure}

A custom-built quadcopter was used during the experiments, as shown in \figref{fig:drone_image}. 
\mmod{It has a mass of 1.5 kg,} and the distance between two diagonal motors is 500 mm. The diameter of each propeller is 254 mm (10 inches). 
The vehicle is equipped with a forward-looking Intel Realsense D455 depth camera for collision avoidance and VIO (with \mmod{the} structure light turned off).
\mmod{The VIO (openVINS \cite{openvins} in the experiments) uses monocular images from the left and right cameras on the D455 stereo depth camera at 15Hz and IMU data at 400Hz to give the estimated states of the vehicle.
It extracts the visual features from the monocular images instead of from the depth images.}
The RAPPIDS planner uses depth images (15Hz) to generate collision-free trajectories.
\par
The proposed perception-aware planner and the VIO run on a small onboard computer (Qualcomm RB5).
The trajectories generated by the perception-aware planner are sent to the Pixracer flight controller running the standard PX4 firmware \cite{px4_firmware}, and are tracked by the low-level position and attitude controllers run on the flight controller. 
\mmod{
The vehicle detects if it reaches the goal point according to the state estimation from VIO -- the goal is reached when the distance between the estimated position of the vehicle and the goal point is below a user-defined threshold.}

\par
\mmod{
Note that the initial heading direction of the vehicle did not need to face  the target position, since the planned trajectory would automatically adjust the yaw angle to always face to the goal during the flight. 
As a result, the ability of the planner to reach the goal is not affected by the initial yaw angle.
We chose the yaw angle to always face the goal to better detect obstacles and free space, as we mentioned in \secref{sec:traj-sampling}.
}

\subsection{Indoor experiments}
\label{sec:indoor_exp}
\begin{figure}[!htp]
    \begin{center}
    \includegraphics[width=\columnwidth]{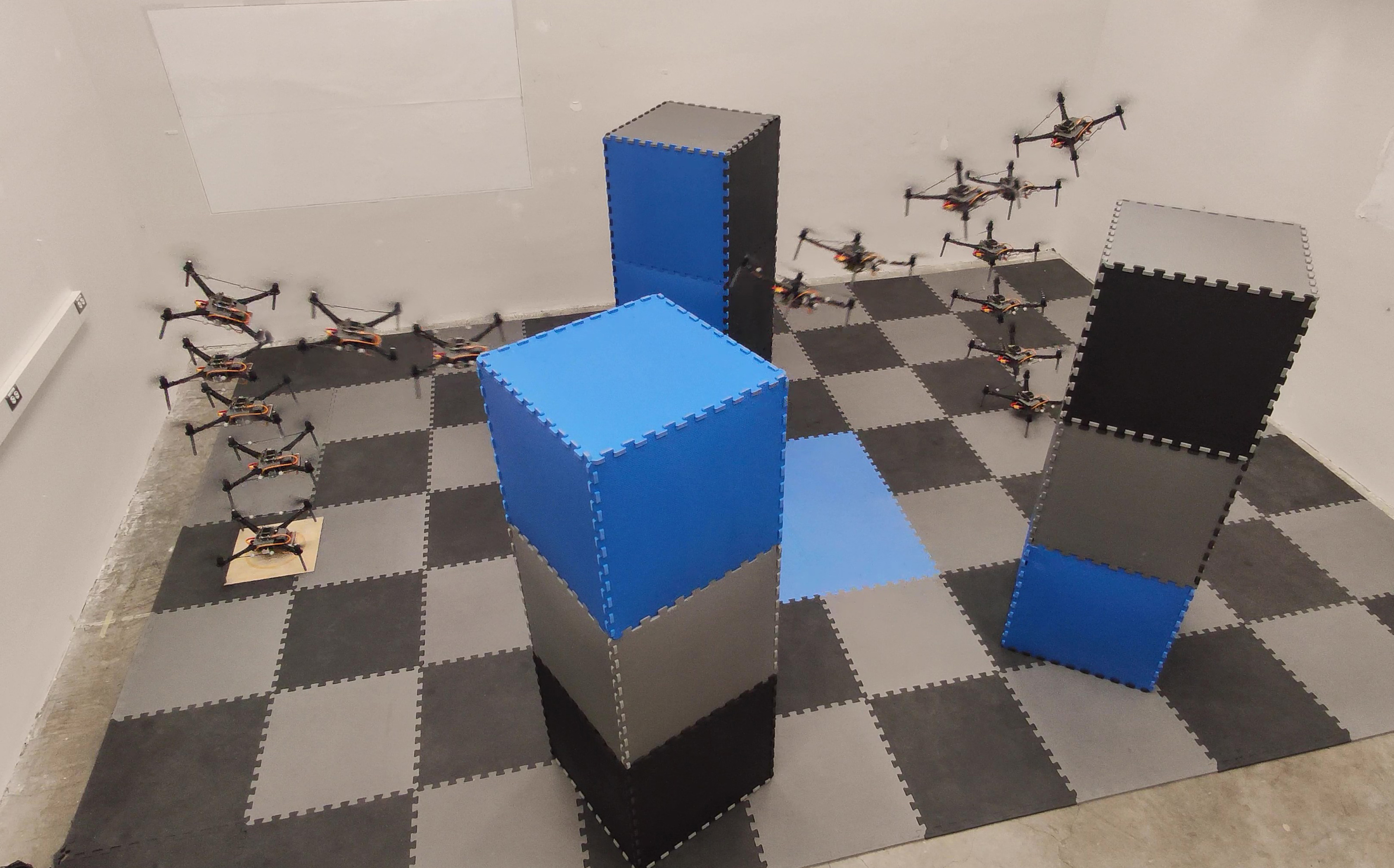}
    \end{center}
    \caption{The indoor experiment setup. 
    The vehicle's goal is 4 meters ahead of the starting point and the vehicle flies from left to right in the image.
    The proposed perception aware planner guides the vehicle to reach the goal avoiding the obstacles on its way, while ensuring good state estimation quality from the VIO.}
    \label{fig:indoor_exp}
\end{figure}

In indoor experiments, we used a motion capture system to provide the ground truth for the vehicle's position. 
As shown in \figref{fig:indoor_exp}, the vehicle first took off to 1.2 meters in height, flew to the target point 4 meters forward avoiding the obstacles, and then landed.
The experiments were repeated for 12 times each for the proposed perception-aware planner and the original RAPPIDS planner, respectively.
The weight of the perception cost $k_{perc}$ was set to 100.
The camera's exposure time $t_\text{exp}$ was set to 8 milliseconds.
Since the planner was only used in the collision avoidance flight, not in the taking-off and landing stages, we exclude the landing stage when comparing the two planners to minimize the uncertainty they introduce.
\par
In the 12 tests, the VIO diverged twice for the original perception-agnostic RAPPIDS planner, while the VIO divergence did not occur for the perception-aware planner. 
Excluding the VIO diverged cases, the performance of the perception-agnostic and perception-aware planners is compared in Table \ref{tab:indoor_exp}.
We can see that, on average, the perception-aware planner reduced the final position estimation error (root mean square error) by 19$\%$. 
Furthermore, the perception-aware planner significantly reduced the aggressiveness of the flights, which can be seen in the reduction of the angular velocity by $24\%$.
This helped reduce the motion blur of the VIO feature points and improved the accuracy of the state estimation.
The flight speed was only slightly slower than the perception-agnostic planner by approximately $6\%$.
In addition, the perception-aware planner kept slightly more feature points within the camera's field of view (6\%). 
The difference in the average feature number was small due to the space-constrained experiment setup, making the geometric paths the vehicle could take similar.

\begin{table}[!htp]
	\renewcommand{\arraystretch}{1.2}
	\caption{Comparison of indoor experimental results between the proposed perception-aware planner and the original perception-agnostic planner.}
	\vspace{-1ex}
	\label{tab:indoor_exp}
	\centering
    \begin{tabular}{|l|c|c|c|}
    \hline
     & original & proposed & difference \\ \hline
    mean final pos. est. error {[}m{]} & 0.141 & 0.114 & -19.1\% \\ \hline
    mean angular velocity {[}rad/s{]} & 1.390 & 1.052 & -24.3\% \\ \hline
    mean feature number in FOV & 97.892 & 103.806 & +6.0\% \\ \hline
    mean speed {[}m/s{]} & 1.376 & 1.298 & -5.7\% \\ \hline
    \end{tabular}
\end{table}

\subsection{Outdoor experiments}
\subsubsection{Environment with distinct visual feature distribution}
\label{sec:outdoor_exp1}
\mmod{The first group of outdoor experiments were conducted along a road near a small forest, where there were much more visual features on the trees than on the road, so that only seeing the road will result in poor VIO accuracy.}
The vehicle first took off to 2 meters in height, flew to the specified target point 20 meters forward, and then landed.
The perception-aware planner guided the vehicle to fly closer to the forest to see more features and at a closer distance to increase the VIO's accuracy, as shown in \figref{fig:outdoor_exp}.
On the contrary, the original perception-agnostic planner flew the vehicle directly to the goal from the starting position.
\par
\begin{table}[!htp]
	\renewcommand{\arraystretch}{1.2}
	\caption{Comparison between planners in an environment with distinct visual feature distribution}
	\vspace{-1ex}
	\label{tab:outdoor_exp1}
	\centering
    \begin{tabular}{|l|c|c|c|}
    \hline
     & original & proposed & difference \\ \hline
    mean pos. est. std [m] & 0.0259 & 0.0213 & -17.8\% \\ \hline
    mean angular velocity {[}rad/s{]} & 0.821 & 0.789 & -3.9\% \\ \hline
    mean feature number in FOV & 45.379 & 50.997 & +12.4\% \\ \hline
    mean speed {[}m/s{]} & 2.419 & 2.225 & -8.0\% \\ \hline
    \end{tabular}
\end{table}

\begin{figure}[!htp]
    \begin{center}
    \includegraphics[width=\columnwidth]{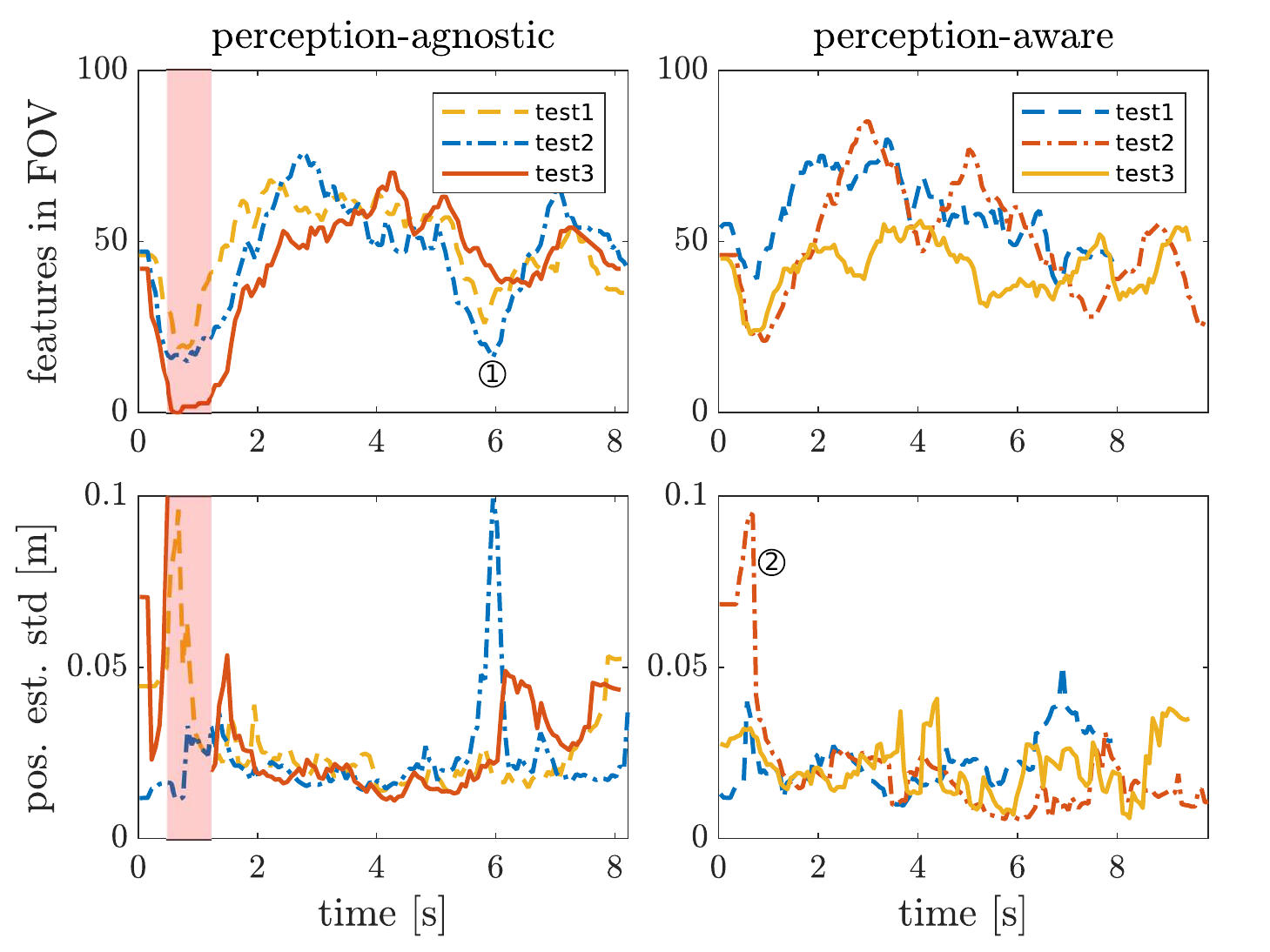}
    \end{center}
    \caption{Comparison between the perception-agnostic planner (column 1) with the perception-aware planner (column 2) \mmod{in the first set of outdoor experiments.} 
    For each planner, the tests are repeated three times, as shown in lines with different color and line type. The red shaded area marks the time when triangulation fails in test 3 of the perception-agnostic planner.
    The proposed perception-aware planner was able to keep more VIO features in the camera's FOV and also reduce position estimation's standard deviation of the vehicle.}
    \label{fig:outdoor_compare1}
\end{figure}

The experiment was repeated three times for each planner (the original perception agnostic planner and the proposed perception-aware planner), and the experimental results are summarized in Table \ref{tab:outdoor_exp1}.
Since both planners were only used in the forward flight, we exclude the taking-off and landing stages.
We can see that the proposed perception-aware planner reduced the standard deviation of position estimation by $17.8 \%$ and increased the number of features in the camera's field of view (FOV) by $12.4 \%$ compared to the original perception-agnostic planner.
The detailed result for each test is shown in \figref{fig:outdoor_compare1}.
In addition, the perception-aware planner prevented the vehicle from seeing very few features in the camera's FOV, which happened at around 1 second in test 3 of the perception-agnostic planner (marked in red in \figref{fig:outdoor_compare1}) and would cause the triangulation to fail.
In test 2 of the perception-agnostic planner, the camera saw a small number of features at around 6 \mmod{seconds} (marked by \raisebox{.5pt}{\textcircled{\raisebox{-.9pt} {1}}}), causing an increase in position estimation uncertainty.
The standard deviation of the position estimation (i.e. \eqref{eq:perc_cost_def} at the current pose of the vehicle, N = 1) is used to evaluate the performance of the VIO instead of the error of position estimation in indoor tests, due to the lack of ground truth from the motion capture system.
Because there was some randomness in the VIO's feature selection, the distribution of features was different at the beginning of the flights, causing the position estimation's standard deviation to be different (marked by \raisebox{.5pt}{\textcircled{\raisebox{-.9pt} {2}}}).
\par

\mmod{The tests were conducted during a sunny day and there was abundant light. 
As a result,} the exposure time $t_\text{exp}$ for the camera was set to 0.05 millisecond, much faster than during the indoor experiments.
This short exposure time meant that the motion blur was much smaller compared to indoors, and the features' motion speed played a very small role in a feature's covariance \eqref{eq:feature_cov}.
As a result, the angular velocity of the perception-aware planner was only $3.9 \%$ lower than that of the perception-agnostic planner, showing similar aggressiveness in flight.
\mmod{Its average flight speed was slightly slower than the perception-agnostic planner, similar to the indoor experiments.}

\subsubsection{Environment with dense obstacles}
\begin{figure}[!tp]
    \centering
    \subfigure[\mmod{Experiment environment with dense obstacles.}]
    {
        \includegraphics[width =0.95\columnwidth]{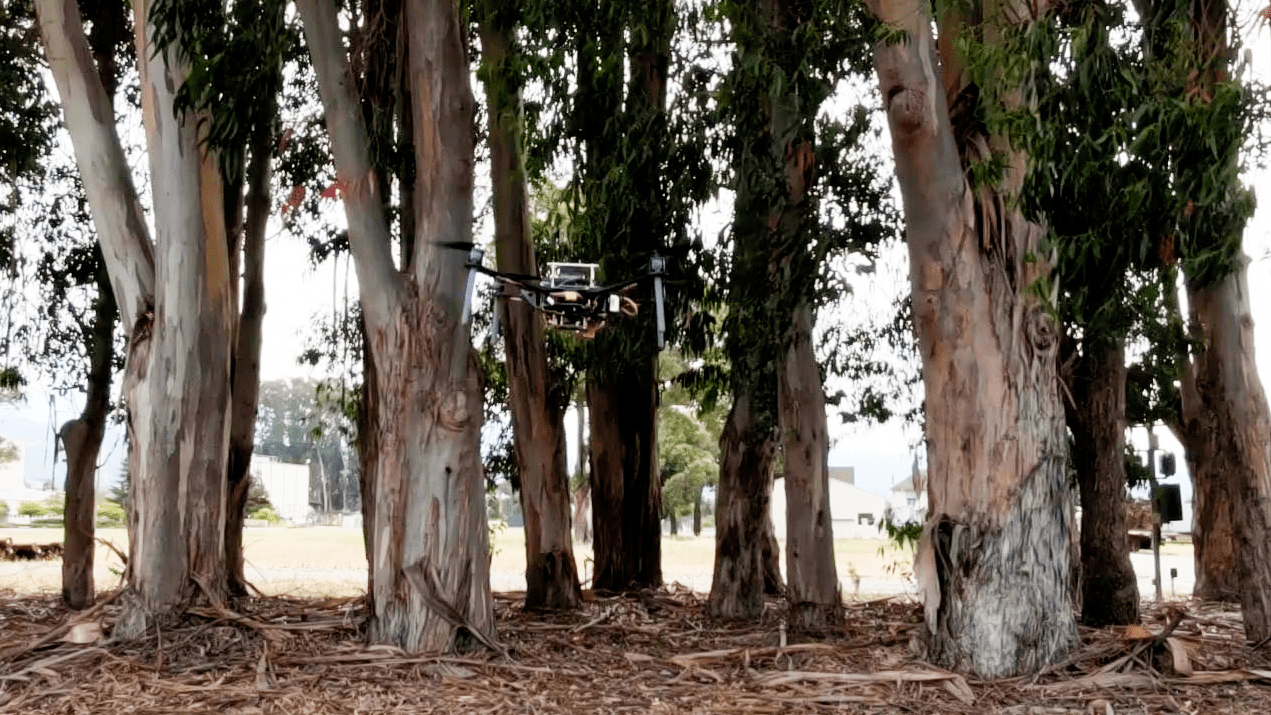}
        \label{subfig:forest-env}
    }
    \subfigure[\mmod{Visualization of the flight path.}]
    {
        \includegraphics[width =0.95\columnwidth]{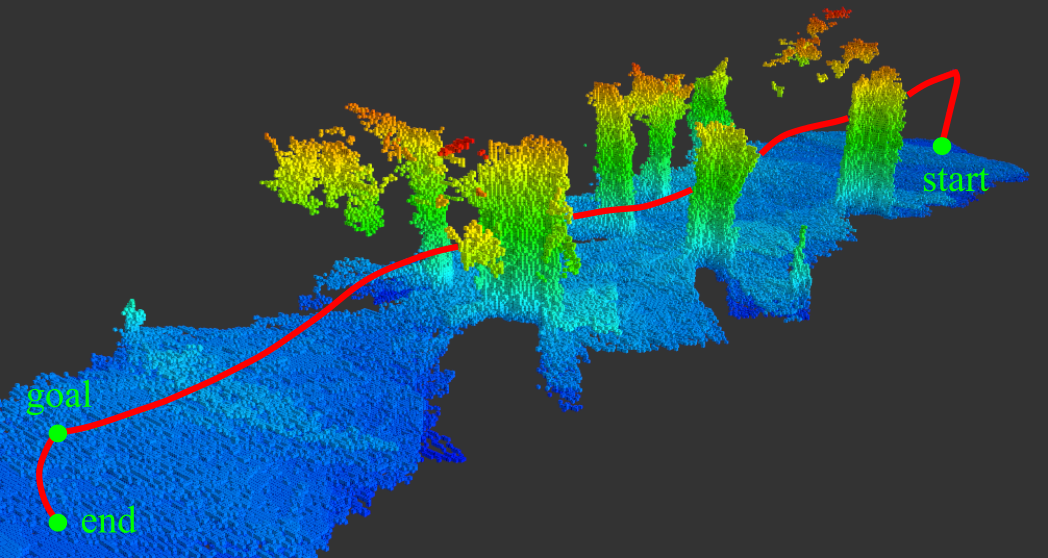}
        \label{subfig:rtabmap-visualization}
    }
    \caption{\mmod{The perception-aware planner guides the vehicle through a small forest, avoiding obstacles while improving the VIO accuracy. (a) The small forest with the vehicle at the beginning of the forward flight. (b) In this sub-figure we visualize the tree obstacles using the RTAB-Map \cite{labbe2019rtab} offline. Our proposed planner didn't use the map during operation. The path of the vehicle is marked in red line, while the start point, goal point and end point are marked with green dots. The distance between the start point and the end point is 30 meters.}}
    \label{fig:outdoor_exp2}
\end{figure}

The second set of outdoor experiments were conducted near a small dense forest, where the vehicle was commanded to take off to 2 meters in height, fly 30 meters forward to pass through the forest, and then land, as shown in \figref{fig:outdoor_exp2}. 
The aim of these experiments is to show the ability of our proposed perception-aware planner to avoid dense obstacles, and simultaneously improve the VIO estimation accuracy.
The tests were conducted during a cloudy day, and the camera's exposure time was set to 0.5 millisecond.

\par
\begin{figure}[!htp]
    \begin{center}
    \includegraphics[width=\columnwidth]{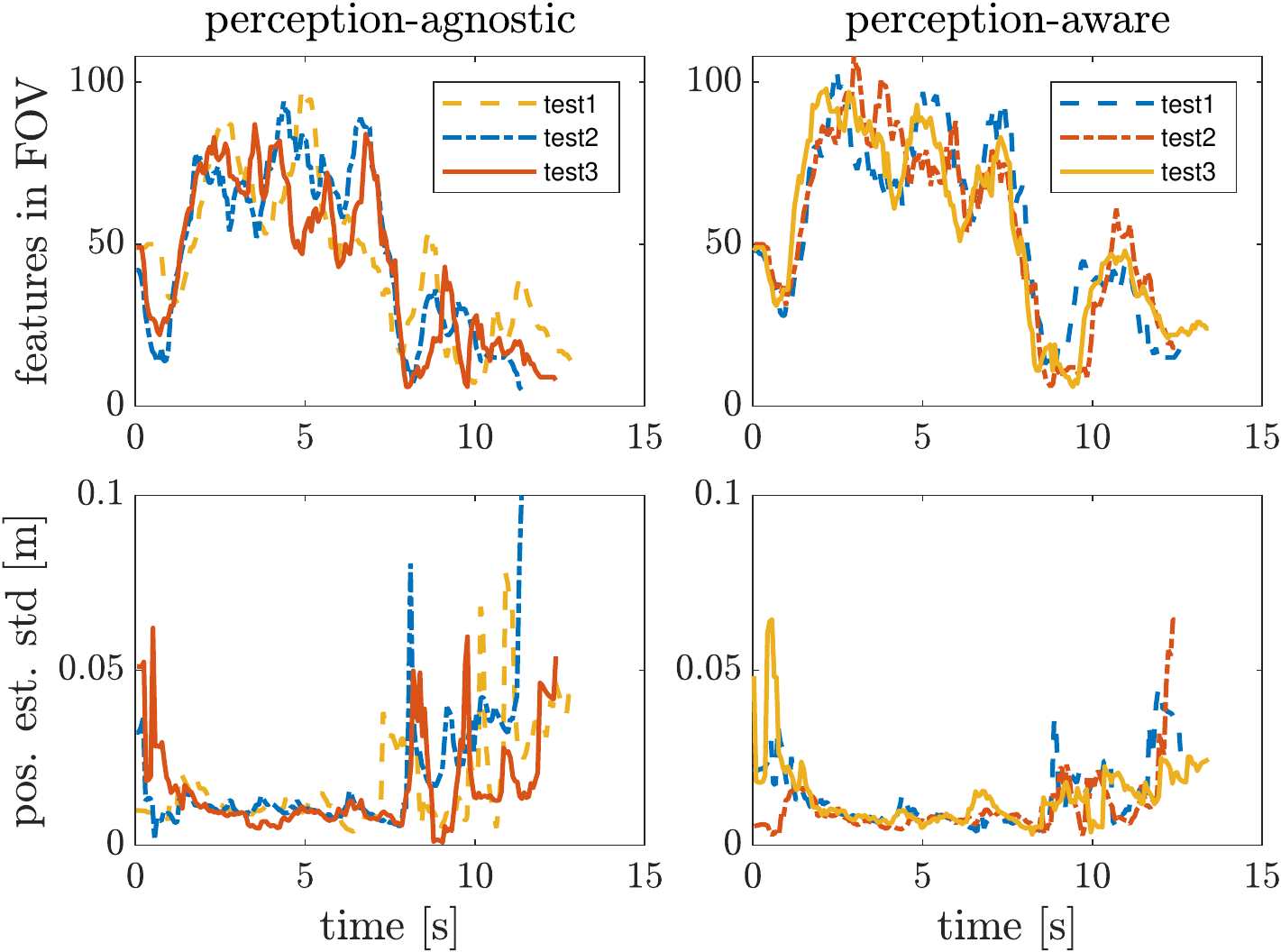}
    \end{center}
    \caption{\mmod{Comparison between the perception-agnostic planner (column 1) with the perception-aware planner (column 2) in the second set of outdoor experiments. For each planner, the tests are repeated three times, as shown in lines with different color and line type.
    Both planner were able to fly through the dense tree obstacles.
    The proposed perception-aware planner was able to keep more VIO features in the camera's FOV and also reduce position estimation's standard deviation of the vehicle.}}
    \label{fig:outdoor_compare2}
\end{figure}

\mmod{
The experiment was repeated three times for each planner (the original perception agnostic planner and the proposed perception-aware planner), and the experimental results are summarized in Table \ref{tab:outdoor_exp2}, like in previous experiments, we exclude the taking-off and landing stages because both planners were only used in the forward flight.
We can see that the perception-aware planner was able to fly the vehicle through the dense forest like the perception agnostic planner.
In addition, it improved the number of visual features in the camera's FOV by $20.6\%$, and also reduced the standard deviation of position estimation by $25.6\%$.
The detailed number of VIO features in the camera's FOV and the standard deviation of position estimation for each of the tests are visualized in \figref{fig:outdoor_compare2}.
We saw in the tests that the perception-aware planner flew the vehicle closer to the forest ground, as there were more features on the ground than on the trees.
This helped the camera to keep more features in the camera's FOV and keep a closer distance to them -- both were helpful for increasing the VIO's accuracy.
The dramatic drop of the number of visual features in the camera's FOV (the first row) at around 8 seconds was because the vehicle flew out of the forest, and a landscape with less visual features (road and grass land) entered the camera's view.

\par
Because the experiments were conducted during a cloudy day, the camera exposure time of 0.5 milliseconds was between the 8 milliseconds exposure time in the indoor experiments (\secref{sec:indoor_exp}) and the 0.05 milliseconds in the first set of outdoor experiments during a sunny day (\secref{sec:outdoor_exp1}).
This led to a degree of motion blur between these two previous experiments for a feature point moving at the same speed in the image plane.
As a result, the reduction in the flight aggressiveness of $14.1\%$, was also in between them ($24.3 \%$ for the indoor tests and $3.9\%$ in the first set of outdoor experiments).
Like in the previous experiments, the introduction of the perception cost  slightly decreased the flight speed, since the aim was not solely fast flight towards the goal as for the perception-agnostic planner.
This led to a $5.56\%$ reduction in flight speed compared with the perception-agnostic planner. 
}
\begin{table}[!htp]
	\renewcommand{\arraystretch}{1.2}
	\caption{\mmod{Comparison between the planners in an environment with dense obstacles}}
	\vspace{-1ex}
	\label{tab:outdoor_exp2}
	\centering
	\mmod{
    \begin{tabular}{|l|c|c|c|}
    \hline
     & original & proposed & difference \\ \hline
    mean pos. est. std [m] & 0.0172 & 0.0128 & -25.6\% \\ \hline
    mean angular velocity {[}rad/s{]} & 0.810 & 0.710 & -14.1\% \\ \hline
    mean feature number in FOV & 45.469 & 54.846 & +20.6\% \\ \hline
    mean speed {[}m/s{]} & 2.392 & 2.259 & -5.56\% \\ \hline
    \end{tabular}
    }
\end{table}

\section{Conclusion and future work}
In this work, we proposed a receding horizon perception-aware local planner for multicopters, which is able to guide the vehicle to areas with rich visual features and reduce the features' motion blur by reducing the planned trajectory's aggressiveness.
\mmod{
By taking the motion blur of the features into consideration, our proposed method is able to adjust the trajectory's aggressiveness based on the light conditions.}
We conducted both indoor and outdoor experiments to show the effectiveness of the proposed method in improving the VIO's position estimation accuracy and reducing the VIO's failure rate.
\mmod{
The ability of the proposed planner to fly through dense obstacles was also demonstrated through the experiments.
Thanks to the computationally efficient collision checking and cost function design,} our method is capable of running in real time on a small embedded computer onboard the vehicle.

\par
\mmod{
A potential future extension of this work is to combine the proposed perception-aware local planner with a global planner for large-scale perception-aware navigation of multicopters in complex environments. 
The global planner can supplement the proposed local planner by giving high-level waypoints and taking into account the semantic information to avoid areas with potential unreliable visual features (like water surface).
Another potential extension is to take dynamic obstacles into account: a module of detecting and predicting the motion of moving obstacles needs to be added.
This module can also be used for preventing the VIO from using the visual features on the moving objects and improve its accuracy.
}

\section*{Acknowledgements} 
This work was partially supported by the AFRI Competitive Grant no. 2020-67021-32855/project accession no. 1024262 from the USDA National Institute of Food and Agriculture.
The AFRI Competitive Grant is being administered through AIFS: the AI Institute for Next Generation Food Systems, \url{https://aifs.ucdavis.edu}.
Research was also partially sponsored by the Army Research Laboratory and was accomplished under Cooperative Agreement Number W911NF-20-2-0105. 
The views and conclusions contained in this document are those of the authors and should not be interpreted as representing the official policies, either expressed or implied, of the Army Research Laboratory or the U.S. Government. 
The U.S. Government is authorized to reproduce and distribute reprints for government purposes notwithstanding any copyright notation herein.
\par
The authors would thank Patrick Geneva of the Robot Perception and Navigation Group, University of Delaware, for his helpful support and suggestions over the usage of the OpenVINS.
The experimental testbed at the HiPeRLab is the result of contributions of many people, a full list of which can be found at \url{hiperlab.berkeley.edu/members/}.

{
\bibliographystyle{IEEEtran}
\bibliography{bib/bibliography}
}

\newpage
\begin{IEEEbiography}[{\includegraphics[width=1in,height=1.25in,clip,keepaspectratio]{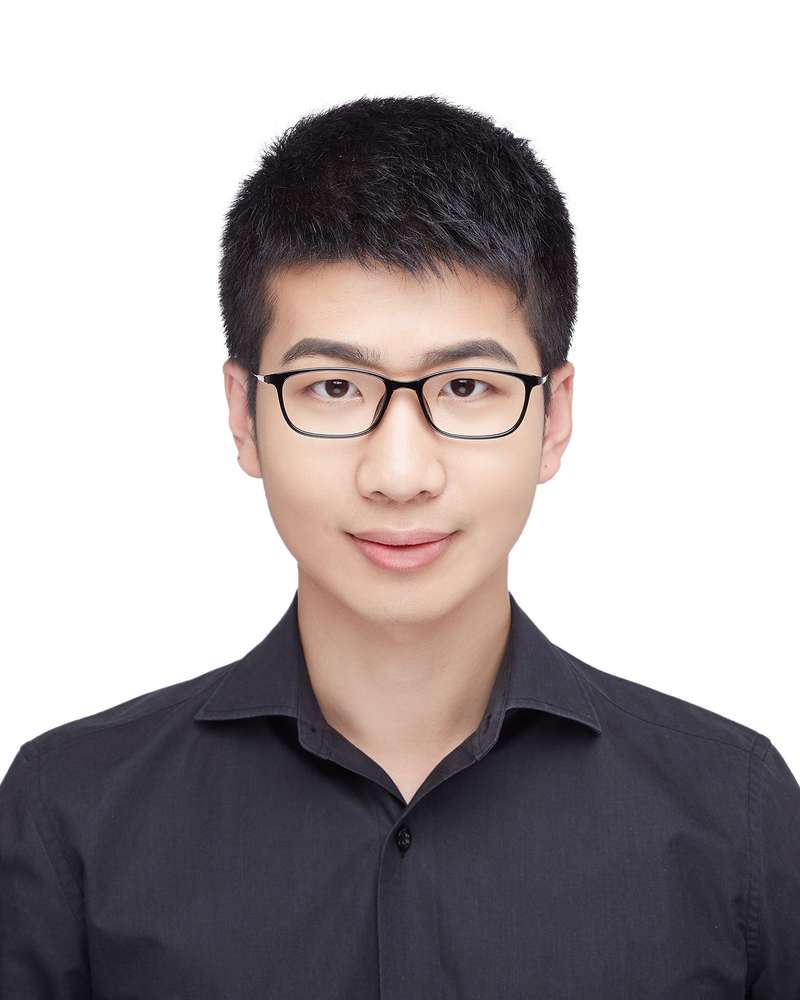}}]{XIANGYU WU} received his bachelor of science degree from Beijing Institute of Technology, China in 2017 and master of science degree from University of California, Berkeley, USA in 2019.
He is currently a Ph.D. candidate at the High Performance Robotics Lab at UC Berkeley.
His current research interests are the state estimation and path planning of multicopters.
\end{IEEEbiography}

\begin{IEEEbiography}[{\includegraphics[width=1in,height=1.25in,clip,keepaspectratio]{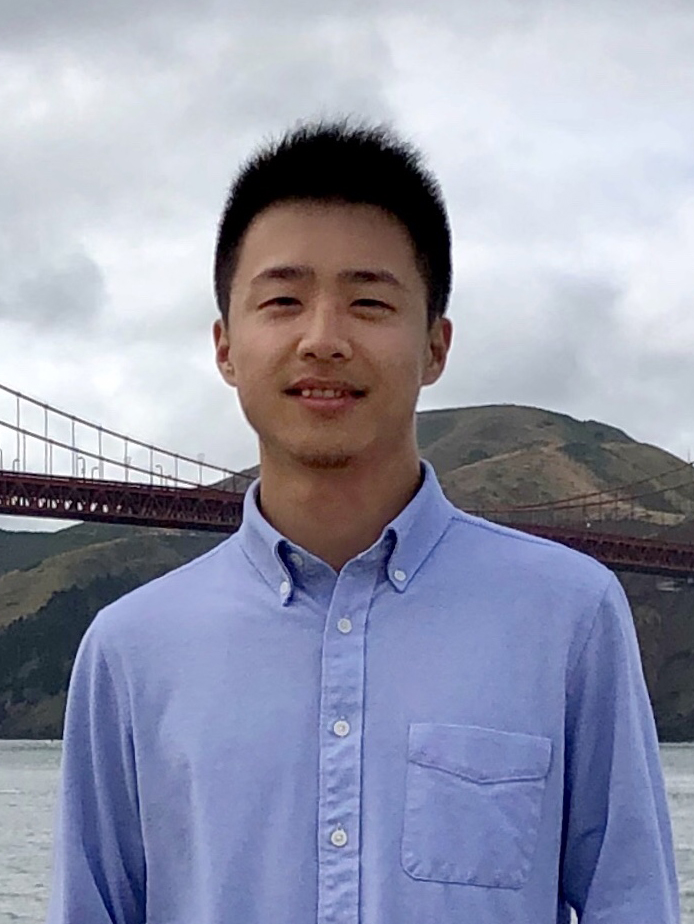}}]{SHUXIAO CHEN} received his bachelor of engineering degree from the University of Nottingham, UK in 2017 and master of engineering degree from the University of California, Berkeley, USA in 2019.
He is currently a Ph.D.~student at the Hybrid Robotics Group at UC Berkeley.
His current research interests are the perception and state estimation for legged robots.
\end{IEEEbiography}

\begin{IEEEbiography}[{\includegraphics[width=1in,height=1.25in,clip,keepaspectratio]{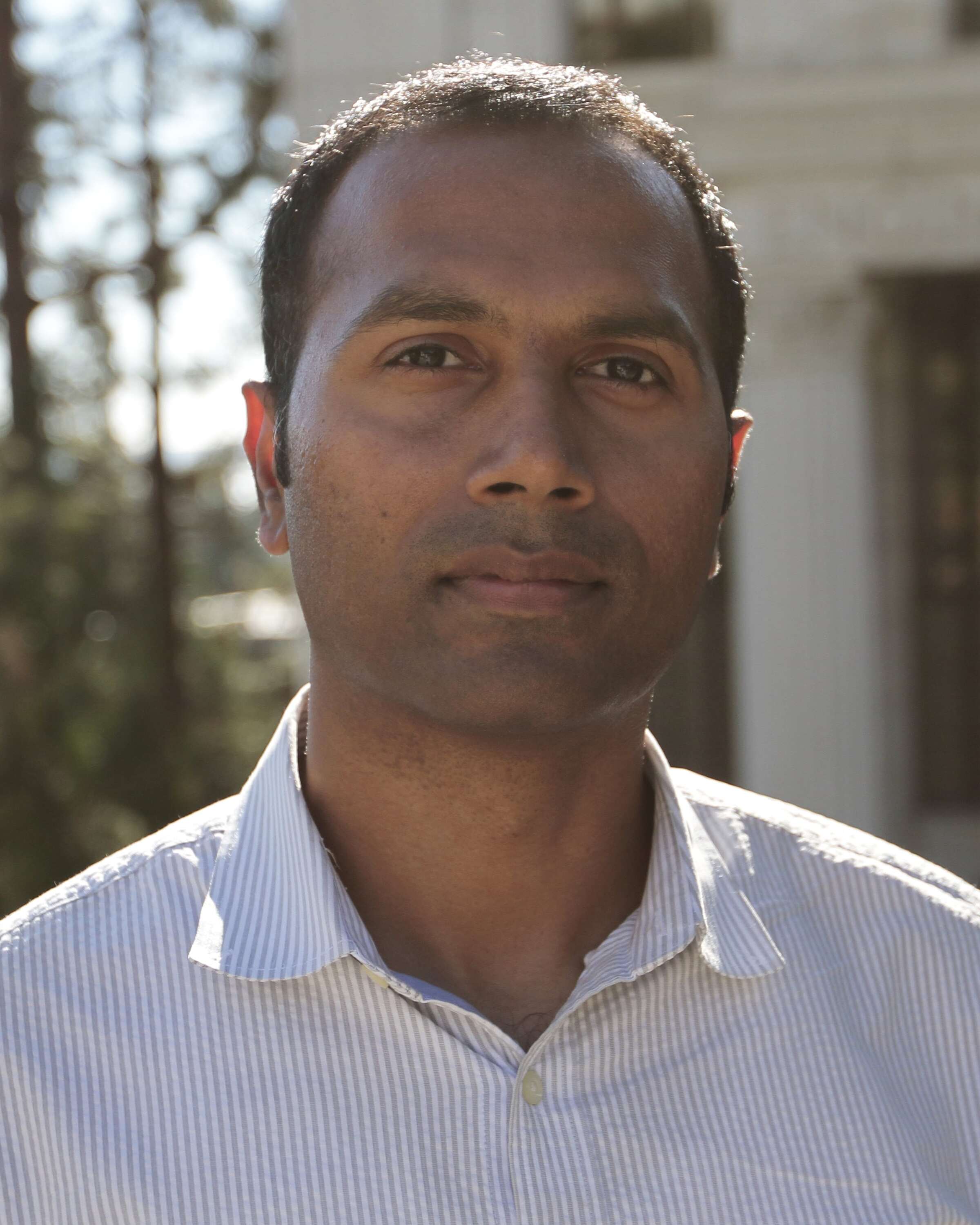}}]{KOUSHIL SREENATH} is an Associate Professor of Mechanical Engineering, at UC Berkeley. He received the Ph.D. degree in Electrical Engineering and Computer Science from the University of Michigan at Ann Arbor, MI, in 2011. He was a Postdoctoral Scholar at the GRASP Lab at University of Pennsylvania from 2011 to 2013 and an Assistant Professor at Carnegie Mellon University from 2013 to 2017. His research interests center on dynamic robotics, applied nonlinear control, and safety-critical control. He received the NSF CAREER, Hellman Fellow, Best Paper Award at the Robotics: Science and Systems (RSS), and the Google Faculty Research Award in Robotics.
\end{IEEEbiography}

\begin{IEEEbiography}[{\includegraphics[width=1in,height=1.25in,clip,keepaspectratio]{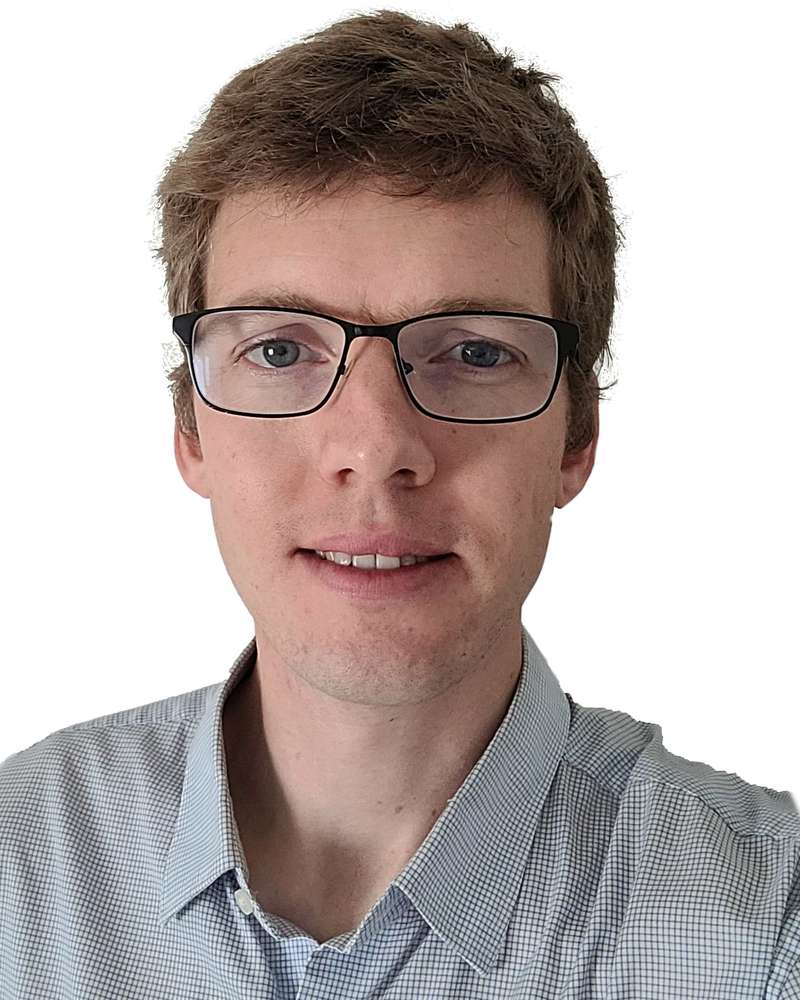}}]{MARK W.\ MUELLER} is an assistant professor of Mechanical Engineering at the University of California, Berkeley, and runs the High Performance Robotics Laboratory (HiPeRLab). He received a Dr.Sc. and M.Sc. from the ETH Zurich in 2015 and 2011, respectively, and a BSc from the University of Pretoria in 2008. His research interests include aerial robotics, their design and control, and especially the interactions between physical design and algorithms.
\end{IEEEbiography}

\end{document}